\documentclass[10pt,journal,compsoc]{IEEEtran}
\usepackage{amsmath,amsfonts}
\usepackage{algorithmic}
\usepackage{algorithm}
\usepackage{array}
\usepackage[caption=false,font=normalsize,labelfont=sf,textfont=sf]{subfig}
\usepackage{textcomp}
\usepackage{stfloats}
\usepackage{url}
\usepackage{verbatim}
\usepackage{graphicx} 
\usepackage{hyperref}
\usepackage{cite}
\usepackage{xcolor}
\usepackage{amssymb}
\usepackage{booktabs}
\usepackage{color}
\usepackage{colortbl}
\usepackage{bbm}
\usepackage{multirow}
\definecolor{origin}{rgb}{0.2,0.3,0.9}
\definecolor{mygray}{gray}{.9}
\definecolor{mygray2}{gray}{.8}
\usepackage{hyperref}
\usepackage{url}
\usepackage{graphicx}
\usepackage{amsmath}
\usepackage{mathrsfs}
\usepackage{wrapfig}
\usepackage{booktabs}       % professional-quality tables
\usepackage{amsfonts}       % blackboard math symbols
\usepackage{nicefrac}       % compact symbols for 1/2, etc.
\usepackage{microtype}      % microtypography
\usepackage{multirow}
\usepackage{multicol}
\usepackage{makecell}
\usepackage{mathtools}
\usepackage{xspace}
\usepackage{amssymb}
\usepackage{algorithm}
\usepackage{algorithmic}
\usepackage{colortbl}
\usepackage{arydshln}
\usepackage{float}
\usepackage[dvipsnames]{xcolor}

\definecolor{citecolor}{HTML}{0071BC}
\definecolor{linkcolor}{HTML}{ED1C24}
\definecolor{grey}{HTML}{999999}
\definecolor{green}{HTML}{ABD1BC}
\definecolor{lightblue}{HTML}{B0C4DE}
\definecolor{purple}{HTML}{E3BBED}
\definecolor{orange}{HTML}{ffdab9}

\definecolor{lightmauve}{rgb}{0.86, 0.82, 1.0}

\begin{document}

\title{MC\#: Mixture Compressor for Mixture-of-Experts Large Models}
% \title{MC\#: Mixture Compressor for Extreme Quantization and Dynamic Experts for  Mixture-of-Experts Vision-Language Models}

\author{
Wei Huang, Yue Liao, Yukang Chen, Jianhui Liu, Haoru Tan,
Si Liu~\IEEEmembership{Senior Member,~IEEE}, \\ Shiming Zhang~\IEEEmembership{Member,~IEEE}, Shuicheng Yan~\IEEEmembership{Fellow,~IEEE}, Xiaojuan Qi~\IEEEmembership{Senior Member,~IEEE}
    % <-this % stops a space
    \IEEEcompsocitemizethanks{
\IEEEcompsocthanksitem Wei Huang, Jianhui Liu, Haoru Tan, Shiming Zhang and Xiaojuan Qi are with the Department of Electrical and Electronic Engineering, The University of Hong Kong, Hong Kong, Hong Kong SAR. Email: weih@connect.hku.hk, jhliu0212@gmail.com, hrtan@eee.hku.hk, beszhang@hku.hk, xjqi@hku.hk
\IEEEcompsocthanksitem Yue Liao and Shuicheng Yan are with the School of Computing, National University of Singapore, Singapore. Email: liaoyue.ai@gmail.com, yansc@nus.edu.sg
\IEEEcompsocthanksitem Yukang Chen is with NVIDIA Research, USA. Email: chenyukang2020@gmail.com
\IEEEcompsocthanksitem Si Liu is with the Institute of Artificial Intelligence, Beihang University, Beijing, China, and also with Hangzhou Innovation Institute, Beihang University, Hangzhou, China. Email: liusi@buaa.edu.cn. 
\IEEEcompsocthanksitem A preliminary version of this research has appeared in ICLR 2025~\cite{huang2024mc}.
    }   
    }

\markboth{}%
{Shell \MakeLowercase{\textit{et al.}}: A Sample Article Using IEEEtran.cls for IEEE Journals}

\maketitle

\begin{abstract}

Mixture-of-Experts (MoE) has emerged as an effective and efficient scaling mechanism for large language models (LLMs) and vision-language models (VLMs). By expanding a single feed-forward network into multiple expert branches, MoE increases model capacity while maintaining efficiency through sparse activation. However, despite this sparsity, the need to preload all experts into memory and activate multiple experts per input introduces significant computational and memory overhead. The expert module becomes the dominant contributor to model size and inference cost, posing a major challenge for deployment. To address this, we propose MC\# (Mixture-Compressor-sharp), a unified framework that combines static quantization and dynamic expert pruning by leveraging the significance of both experts and tokens to achieve aggressive compression of MoE-LLMs/VLMs. To reduce storage and loading overhead, we introduce Pre-Loading Mixed-Precision Quantization (PMQ), which formulates adaptive bit allocation as a linear programming problem. The objective function jointly considers expert importance and quantization error, producing a Pareto-optimal trade-off between model size and performance. To reduce runtime computation, we further introduce Online Top-any Pruning (OTP), which models expert activation per token as a learnable distribution via Gumbel-Softmax sampling. During inference, OTP dynamically selects a subset of experts for each token, allowing fine-grained control over activation. By combining PMQ’s static bit-width optimization with OTP’s dynamic routing, MC\# achieves extreme compression with minimal accuracy degradation. On DeepSeek-VL2, MC\# achieves a 6.2× weight reduction at an average of 2.57 bits, with only a 1.7\% drop across five multimodal benchmarks compared to the 16-bit baseline. Moreover, OTP further reduces expert activation by 20\% with less than 1\% performance loss, demonstrating strong potential for efficient deployment of MoE-based models.

\end{abstract}


\section{Introduction}
\begin{figure}[!t]
% \vspace{-0.2in}
\centerline{\includegraphics[width=0.48\textwidth]{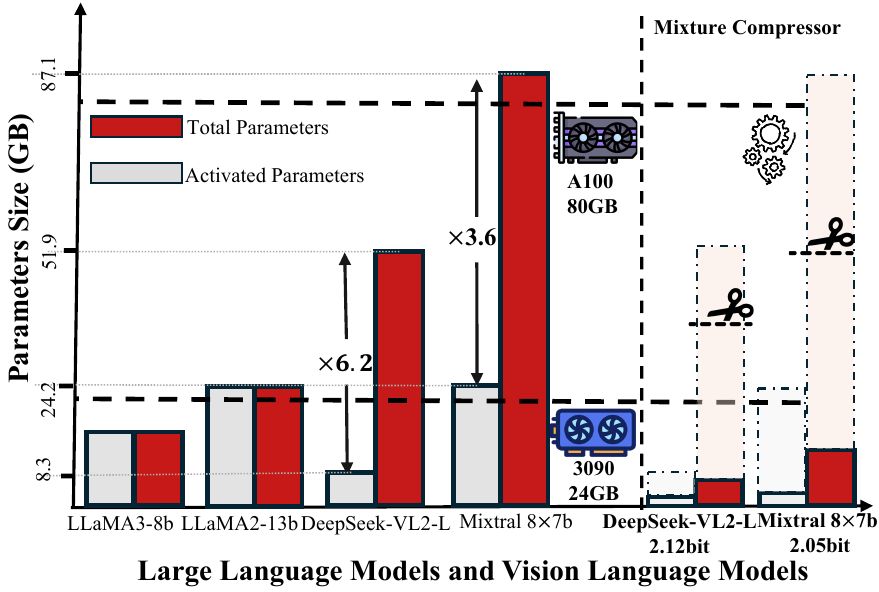}}

\caption{Comparison of total parameter size and inference activated parameter size on a few open-source large vision/language models and compressed Mixtral 8$\times$7b (MoE-LLMs) and DeepSeek-VL2-L (MoE-VLMs). L: large.}
\label{fig:1}

\end{figure}

%introduce moe, but heavy to deploy and running 
% Mixture-of-Experts large language models and vision-language models (MoE-LLMs/VLMs)~\cite{muennighoff2024olmoe,jiang2024mixtral,dai2024deepseekmoe} provide an efficient model-scaling mechanism by utilizing a sparse architecture, in which only a subset of experts is activated by router. This selective activation boosts computational efficiency and scalability by assigning experts dynamically based on the specific needs of each input. Despite reducing the number of active experts to improve inference efficiency, MoE models still face significant deployment challenges. All experts must be loaded into memory simultaneously, and typically at least two experts are activated during inference, resulting in considerable memory and computational overhead.  Even an NVIDIA A100-80GB GPU cannot accommodate typical MoE models like Mixtral 8$\times$7b~\cite{jiang2024mixtral} (Fig.~\ref{fig:1}(b)). The proposed challenges hinder the deployment of LLM with limited hardware resources which further promotes study on MoE-LLM compression for better deploying  model-scaling paradigm.
Mixture-of-Experts large-language/vision-language models (MoE-LLMs/VLMs)~\cite{muennighoff2024olmoe, jiang2024mixtral, wu2024deepseek, lin2024moe} provide an efficient mechanism for model scaling by leveraging sparse architectures, where the router activates only a subset of experts. This selective activation improves computational efficiency and scalability by dynamically assigning experts based on the specific requirements of each input. However, while the number of active experts is reduced to enhance inference efficiency, all experts must still be loaded into memory simultaneously, and multiple experts are typically activated during inference. This results in substantial memory and computational overhead. For instance, a typical MoE-LLM like Mixtral 8$\times$7b~\cite{jiang2024mixtral} requires nearly 90GB of GPU memory (Fig.~\ref{fig:1}), while MoE-VLMs such as DeepSeek-VL2-L demand over 50GB of GPU memory (Fig.~\ref{fig:1}), largely due to the heavy computational and memory requirements of their LLM backbone. These high resource requirements hinder the deployment of MoE-based large models on hardware with limited capacity, further driving the need for research into compression techniques for MoE large models.

% The primary goal of compressing MoE large models is to reduce the size of expert parameters, as they dominate the memory usage~\cite{li2024examining, huang2024mc}. For instance, in models like Mixtral $8\times7$b, the number of expert parameters is 33 times greater than that of the attention modules. On the other hand, recent studies ~\cite{chi2022representation,lu2024not} have shown that due to the training strategies of MoE, not all experts are equally important, which indicates that both the static experts during the pre-loading phase and the dynamic experts during online inference need to be compressed. Previous expert compression methods have typically focused on compressing a single phase, such as quantizing expert weights during the pre-loading stage ~\cite{li2024examining} or pruning experts during the inference stage ~\cite{lu2024not,koishekenov2022memory,kim2021scalable}. Furthermore, vanilla uniform bit-width quantization and expert pruning based solely on routing scores struggle to maintain performance at extremely high compression ratios. Therefore, in this work, we are the first to explore extreme training-free mixture compression for MoE-LLMs, efficiently combining static expert quantization with dynamic expert pruning using a combination of expert importance metrics to achieve ultra-lightweight MoE-LLMs without significantly sacrificing performance.
The primary objective of compressing MoE-based large models is to reduce the size of expert parameters, as they significantly impact memory usage~\cite{li2024examining, huang2024mc}. For example, in models such as DeepSeek-VL2-L, the number of expert parameters is about 77 times larger than that of attention modules. On the other hand, recent studies~\cite{chi2022representation, lu2024not, huang2024mc} have revealed that, due to the training strategies of MoE models, not all experts are equally important, and the importance of input tokens also varies. This suggests that compressing MoE models can benefit from quantizing expert weights during the preloading phase~\cite{li2024examining} or pruning experts during inference~\cite{lu2024not, koishekenov2022memory, kim2021scalable}. However, traditional uniform bit-width quantization struggles to maintain performance under extreme compression ratios, and rule-based expert pruning is often unsuitable for MoE models with a large number of experts, such as DeepSeek-VL2. To address these challenges, this study explores extreme hybrid compression for multimodal MoE models. By introducing expert importance metrics, we apply mixed-precision quantization to statically compress experts. Additionally, we propose an innovative differentiable mask based on Gumbel-Softmax sampling, enabling dynamic top-any expert pruning for different tokens. This approach achieves highly lightweight MoE-VLMs while maintaining an optimal Pareto curve between compression and performance.

\begin{figure*}[!t]
% \vspace{-0.2in}
\centerline{\includegraphics[width=1\textwidth]{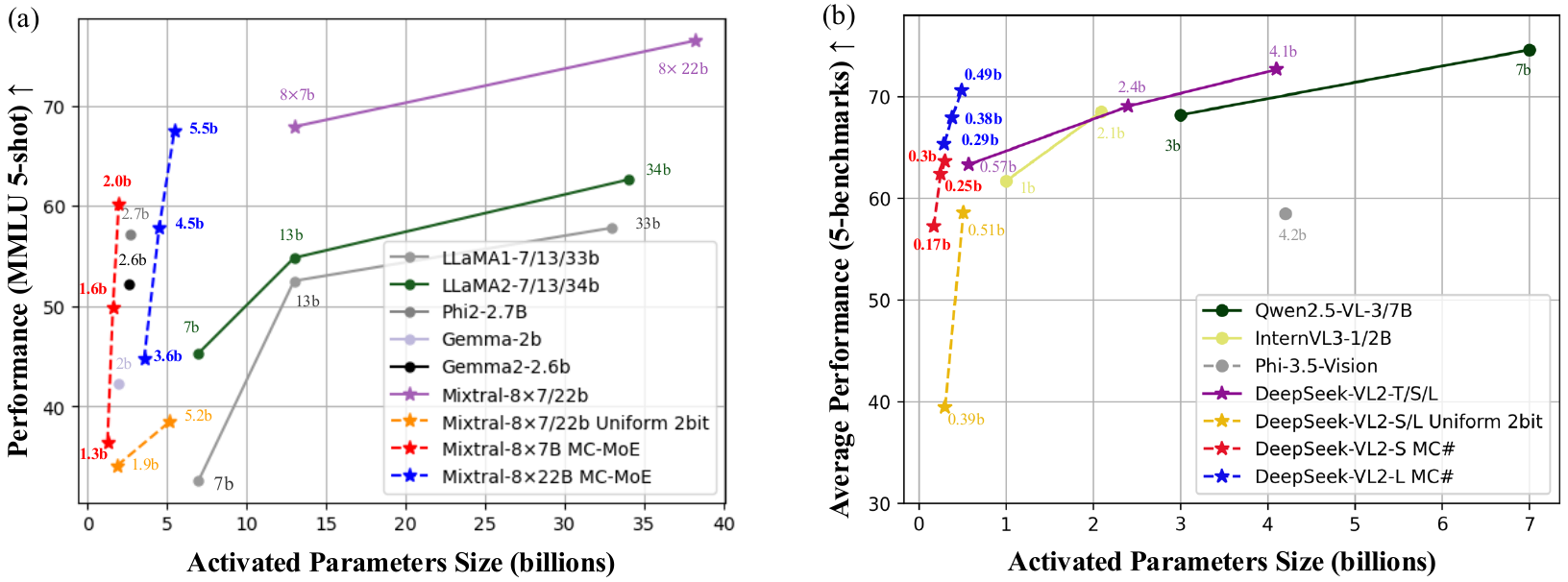}}
% \vspace{-0.1in}
\caption{(a) MMLU (5-shot$\uparrow$) accuracy across different open-source LLMs with various activated parameters (dot-lines denote the quantized models, solid-lines are 16-bit models). To align quantized models' parameter size with 16-bit models, we define 16bits as one standard parameter (e.g., 8$\times$2-bit elements represent one parameter). (b) Average performance($\uparrow$) on 5 general multimodal benchmarks across different open-source VLMs with various activated parameters. L: large, S: small. T: tiny.}
\label{fig:2}

\end{figure*}

\begin{figure*}[!t]
% \vspace{-0.2in}
\centerline{\includegraphics[width=1\textwidth]{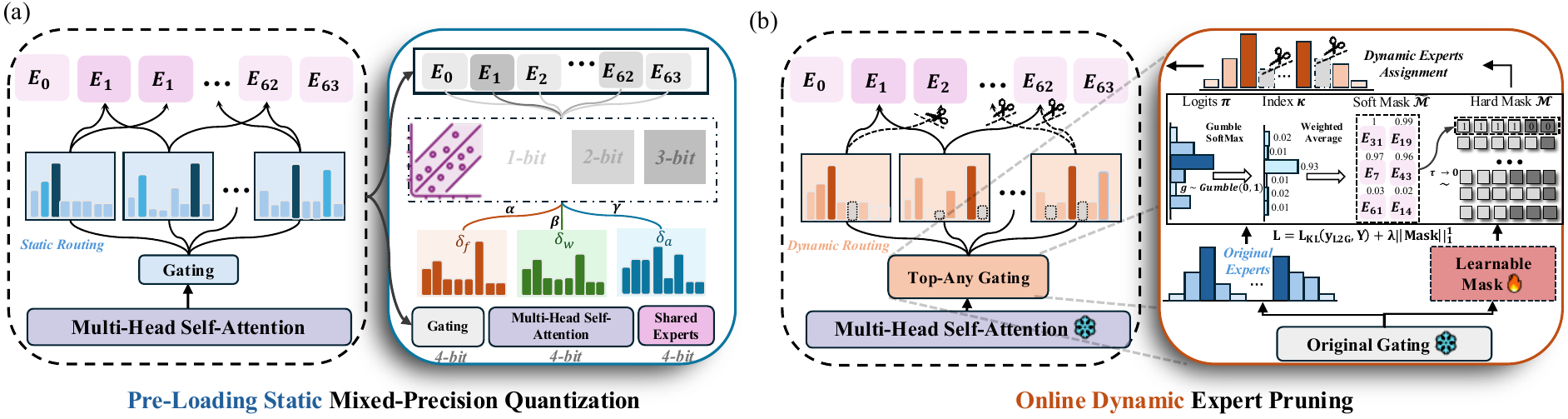}}
\caption{The overview of our proposed MC pipeline with two-stage compression for experts. (a) Framework of pre-loading static mixed-precision quantization (PMQ) of MoE-LLMs. PMQ determines the activated feature and loss sensitivity of all experts and plans the optimal precision configuration under ultra-low-bit-width. (b)  Schematic of online top-any pruning (OTP) of MoE-LLMs. OTP utilizes a learnable experts pruning scheme to achieve higher inference efficiency.}
\label{fig:3}
\end{figure*}

We introduce \textbf{MC\#}, \textit{i.e.}, \textbf{M}ixture-\textbf{C}ompressor-sharp, a novel framework for multimodal MoE models that integrates expert quantization and pruning for efficient compression. \textbf{MC\#} operates in two key phases: \emph{Pre-Loading Mixed-Precision Quantization~(PMQ)} and \emph{Online Top-any Pruning
(OTP)}, as illustrated in Fig.~\ref{fig:3}. 

In the pre-loading phase, we achieve extreme compression of expert parameters through mixed-precision quantization under ultra-low bits. Analysis of MoE-LLMs reveals significant imbalances in activation reconstruction errors, routing weights, and expert activation frequencies~\cite{huang2024mc}, with these discrepancies becoming more pronounced in MoE-VLMs (see Section~\ref{sec:2.2.1} and Fig.~\ref{fig:4}). These imbalances motivate assigning distinct bit-widths to different experts in VLMs. We propose a weighted evaluation function that jointly accounts for activation frequency, activation weight, and quantization losses across bit-widths. This function is optimized using a Linear Programming (LP) model to determine the optimal quantization configuration. Utilizing a training-free Post-Training Quantization (PTQ) approach, such as GPTQ~\cite{frantar2022gptq}, our PMQ framework delivers high-performance compression at any-precision in low bit-widths (1.5-bit to 2.5-bit) while striking a Pareto-optimal balance between model size and performance. Furthermore, our mixed-precision strategy seamlessly integrates with state-of-the-art quantization techniques~\cite{tseng2024quip, chen2024efficientqat, shao2023omniquant, egiazarian2024extreme, liao2024apiq}.
% In the pre-loading phase, we focus on extreme compression of the stored experts through low-bit quantization. Our empirical study reveals imbalances in activation reconstruction error, routing weights, and frequencies of activated expert (Sec.~\ref{sec:2.2.1} and Fig.~\ref{fig:4}), which inspires the allocation of different bit-widths to each expert. However, relying solely on the routing frequencies or scores is insufficient to accurately determine the optimal bit-width, as the two distributions may not be consistent but rather the opposite~\cite{li2024examining}. Therefore, we developed a weighted evaluation function that considers both the frequency and scores of expert activations, as well as the associated quantization loss at different bit-widths. This function is then minimized within a Linear Programming (LP) model to determine the optimal quantization configuration. Utilizing a training-free Post-training Quantization~(PTQ) approach, GPTQ~\cite{frantar2022gptq}, \emph{PMQ} achieves high-performance compression at extremely low bit-widths (1.5-bit$\sim$2.5-bit), and our mixed-precision strategy is compatible with other advanced quantization techniques~\cite{tseng2024quip,chen2024efficientqat,shao2023omniquant,egiazarian2024extreme,liao2024apiq}.
As for the inference phase, to address the inflexibility of previous rule-based expert pruning methods~\cite{huang2024mc,lu2024not}, we propose \emph{Online Top-any Pruning (OTP)}, a learnable dynamic pruning approach that operates in an end-to-end training mode with only a small batch of data. In the $top\mbox{-}k$ MoE mechanism, dynamic pruning of experts requires selective activation based on the varying importance of tokens within a limited set of experts. However, the non-differentiability of mask selection prevents direct backpropagation for learning. To overcome this limitation, we model mask selection as a probabilistic sampling process, transforming it into a stochastic operation. By leveraging Gumbel-Softmax~\cite{jang2016categorical}, we make the sampling process differentiable, enabling the optimization of each mask candidate's probability through gradient descent. During training, OTP aims to learn an appropriate token-wise expert mask distribution, further reducing the number of activated experts and single-device computational costs in ultra-low-bit MoE models, all while maintaining the original performance.

% \emph{ODP} dynamically prunes low-confidence experts for each token based on the routing weights. Our pruning strategy follows two key principles: first, experts with significantly lower routing scores are categorized as {``}low confidence" and can be pruned~\cite{lu2024not}. Second, to prevent attention degradation that solely relies on routing weights, we protect important tokens by considering both attention scores and feature magnitudes. Experiments show that protecting only 2\% of the important tokens effectively mitigates pruning loss while maintaining nearly the same compression ratio.

% The proposed mixture compression of low bit-width experts improves performance compared to uniform quantized experts or other mixed-precision strategies, even surpassing float-point (FP) models with the same number of activated parameters. Moreover, when compressing Mixtral $8\times7$b to around 8b (2.54-bit), its activated parameters amounted to only 2b, while even outperforming 16-bit LLaMA2-13b by around 8\% on the MMLU (5-shot), as shown in Fig.~\ref{fig:1}(a). Mixture compression exploits the disparities between MoE experts, for the first time enabling surpassing of smaller FP models of equivalent size under extreme compression without training. This achievement underscores the significant compression potential and practical utility of sparse MoE-LLMs.

Compared to uniform quantization or other mixed-precision strategies, the proposed \textbf{MC\#} approach significantly enhances the performance of compressed MoE-LLMs/VLMs. Specifically, when compressing DeepSeek-VL2-L to approximately 2.57 bits, its activation parameters are reduced to just 0.49b, while maintaining only 1.7\% performance loss across five common multimodal benchmarks. Remarkably, under the same activation parameter budget, it even outperforms uncompressed state-of-the-art small-scale models, as shown in Fig.~\ref{fig:2}(b). This mixture compression strategy fully exploits the unique characteristics of experts, enabling high performance even under extreme compression conditions. These results highlight the immense potential and practicality of compressing multimodal MoE models, making them scalable from multi-GPU cluster environments to consumer-grade and edge-level applications.

\vspace{0.02in}
% \noindent\textbf{Extension to the conference version:}
This paper presents substantial extensions to the conference version ~\cite{huang2024mc} in the following aspects. (i) We conduct an in-depth analysis of the uneven expert activation patterns and compression challenges in MoE-VLMs, such as DeepSeek-VL2, providing valuable insights for effectively deploying mixed-precision quantization in multimodal MoE models. (ii) We propose a differentiable online dynamic expert mask learner based on Gumbel Softmax, which leverages a soft mask mechanism to learn an optimal token-wise expert pruning strategy even with a small number of samples. Additionally, we introduce expert sparsity control constraints to explore the optimal pruning strategies under specific pruning ratios. Our proposed \emph{OTP} strategy successfully overcomes the limitations of rule-based expert pruning methods, demonstrating robust scalability to scenarios with a larger number of experts. (iii) We comprehensively evaluate the compression performance of \textbf{MC\#} on nine language benchmarks and six multimodal large model benchmarks. By combining static and dynamic compression, our approach achieves the state-of-the-art compression performance than other methods and previous MC~\cite{huang2024mc}. (iv) We have demonstrated that our proposed training-free multi-factor mixed-precision quantization method achieves the Pareto-optimal boundary in the ultra-low bit-width range across various types of large MoE models. Furthermore, when combined with a learnable expert pruning strategy, our approach achieves higher compression ratios with significantly lower performance loss. In summary, these advancements contribute to building a more efficient and unified mixed compression framework for MoE-based LLMs and VLMs, showcasing significant potential and applicability across various real-world scenarios.

\section{Related Works}

\subsection{Mixture-of-Experts Large Models}

Large-scale models built on LLM-based architectures have achieved remarkable progress in various natural language processing~\cite{chang2024survey, zhao2023survey} and multimodal understanding domains~\cite{lin2024moe, lu2024deepseek, wu2024deepseek}. Sparse-activation MoE models have emerged as a key strategy for improving the trade-off between performance and efficiency in LLMs and VLMs. In MoE models, each layer consists of multiple expert submodels, and only a specific subset of experts is activated for each token. This mechanism significantly enhances efficiency compared to dense models that activate all parameters for every input~\cite{shazeer2017outrageously, yun2024toward}, while also delivering stronger representational power and improved multimodal understanding. Recent advancements in LLMs~\cite{guo2025deepseek} have further extended the capabilities of MoE-based architectures. Industry-leading models, such as DeepSeek-VL2~\cite{wu2024deepseek}, leverage MoE-LLM backbones to achieve state-of-the-art performance. Despite their success, these models bring more memory requirements on GPU storage, presenting new challenges for efficient deployment~\cite{zhou2024survey, zhu2023survey}.

% LLMs and VLMs have achieved significant advancements across various natural language processing~\cite{chang2024survey,zhao2023survey} and multimodal understanding~\cite{lin2024moe,lu2024deepseek,wu2024deepseek} domains. Despite their success, these models rely heavily on dense parameters, which presents significant challenges for deployment\cite{zhou2024survey,zhu2023survey}. Sparse activated MoE models have been identified as an essential strategy to enhance the cost-performance balance in LLMs and VLMs. In MoE models, each layer is comprised of several experts, with each token activating only a specific subset, thereby significantly improving efficiency compared to dense models, which activate all parameters for every input~\cite{shazeer2017outrageously, yun2024toward}. Recent advancements in VLMs~\cite{wu2024deepseek} have further popularized MoE-based architectures. Industry-leading models such as Mixtral $8\times7$b~\cite{jiang2024mixtral} and Deepseek-R1~\cite{guo2025deepseek} also incorporate this technology. 

\subsection{Quantization for LLMs}

Post-Training Quantization (PTQ) is an efficient approach that requires no additional training, making it particularly suitable for LLMs, such as GPTQ~\cite{frantar2022gptq}, AWQ~\cite{lin2024awq}, and MBQ~\cite{li2025mbq}, and other activation quantization works~\cite{xiao2023smoothquant, shao2023omniquant}. Previous studies~\cite{dong2020hawq, shang2023pb, huang2024billm} have explored the significance of weight diversity and proposed mixed-precision strategies to improve low-bit performance by assigning different bit-widths accordingly. Examples include unstructured relaxed precision methods like SpQR~\cite{dettmers2023spqr} and structured mixed-precision approaches such as SliM-LLM~\cite{huang2024slim}. Recent work has introduced expert-guided block-wise quantization benchmarks for MoE-LLMs to address disparities in expert weights~\cite{li2024examining}; however, significant performance degradation persists under ultra-low-bit quantization. Codebook-based encoding methods offer more precise quantization for LLMs and can further enhance post-quantization performance through fine-tuning~\cite{egiazarian2024extreme, tseng2024quip}. Although Quantization-Aware Training (QAT) demands substantial resources~\cite{chen2024efficientqat, liu2023llm}, QAT-based retraining strategies, or PTQ combined with additional fine-tuning~\cite{liao2024apiq, guo2023lq, huang2024good}, have proven to be more effective in preserving the performance of lightweight, quantized models. Since the LLM component in MoE-VLMs contains the largest number of parameters and incurs the highest inference cost, these PTQ techniques can also be effectively adapted for use in VLMs~\cite{huang2024good}.

\subsection{Parameter pruning for LLMs}
% Parameter pruning is another effective method for neural network compression~\cite{kwon2022fast,hubara2021accelerated}, and it has recently become crucial in reducing the size of LLM weights~\cite{frantar2023sparsegpt,sun2023simple,fang2024maskllm}. Traditional pruning approaches focus on two main techniques: structured and unstructured pruning, both of which selectively zero out certain parameters based on their importance~\cite{zhou2024survey}. In MoE-LLMs, less important experts can be pruned based on activation frequencies or the statistical characteristics of gating~\cite{kim2021scalable, koishekenov2022memory, liu2024efficient}. During the model preloading stage, pruning tends to incur greater loss than quantization at the same compression rate. However, dynamically adjusting the quantization bit-width during inference remains a challenge, whereas pruning offers the flexibility to dynamically select activation parameters during inference~\cite{zhu2023survey}. Recent work by~\cite{lu2024not} has explored dynamically activating $top\mbox{-}k$ experts based on gating weights in MoE, significantly improving inference efficiency.

Parameter pruning is another effective neural network compression method~\cite{kwon2022fast, hubara2021accelerated}, and it has recently become crucial for reducing the size of LLM weights~\cite{frantar2023sparsegpt, sun2023simple,fang2024maskllm}. Traditional pruning techniques typically focus on two approaches: structured pruning and unstructured pruning, both of which selectively zero out parameters based on their importance~\cite{zhou2024survey}. In the context of large MoE models, less important expert models can be pruned based on activation frequency or gating statistics~\cite{kim2021scalable, koishekenov2022memory, liu2024efficient}. During the model pre-loading stage, pruning often incurs greater performance loss than quantization at the same compression ratio. However, while dynamically adjusting quantization bit-widths during inference remains challenging, pruning offers the flexibility to dynamically select active parameters during inference~\cite{zhu2023survey}. Recent studies, such as~\cite{lu2024not}, have explored dynamic activation of $top\mbox{-}k$ (where $k=1$ or $2$) experts in MoE models based on gating weights, significantly improving inference efficiency. However, such straightforward rule-based dynamic activation strategies lack robustness when scaling to larger $k$, as in models like DeepSeek-VL2 ($k=6$)~\cite{wu2024deepseek}.

\begin{figure*}[!t]
\centerline{\includegraphics[width=1\textwidth]{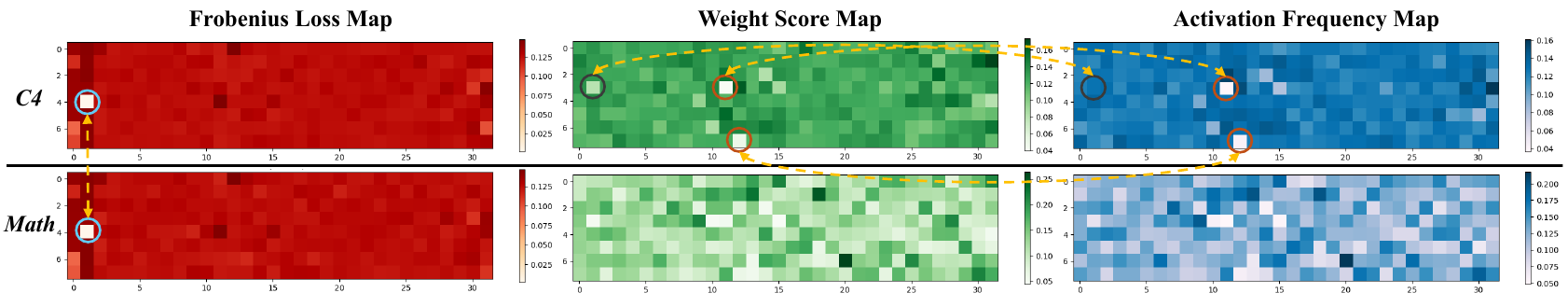}}
\caption{{Distribution of expert drop F-norm (\textcolor{red}{red}), activated weights (\textcolor{ForestGreen}{green}) and frequencies (\textcolor{RoyalBlue}{blue}) in the Mixtral $8\times7$b model, encompassing 32
MoE layers with 8 experts per layer. The top set of the heatmap is calculated through C4 dataset~\cite{raffel2020exploring}, and the bottom set is calculated through MATH dataset. MoE-LLMs selectively activate top$\mbox{-}$2 experts in each MoE layer, wherein a significant portion of experts remain less important or inactivated all the time.}}
% \vspace{-0.25in}
\label{fig:4}
\end{figure*}

\begin{figure*}[h]
\centerline{\includegraphics[width=1\textwidth]{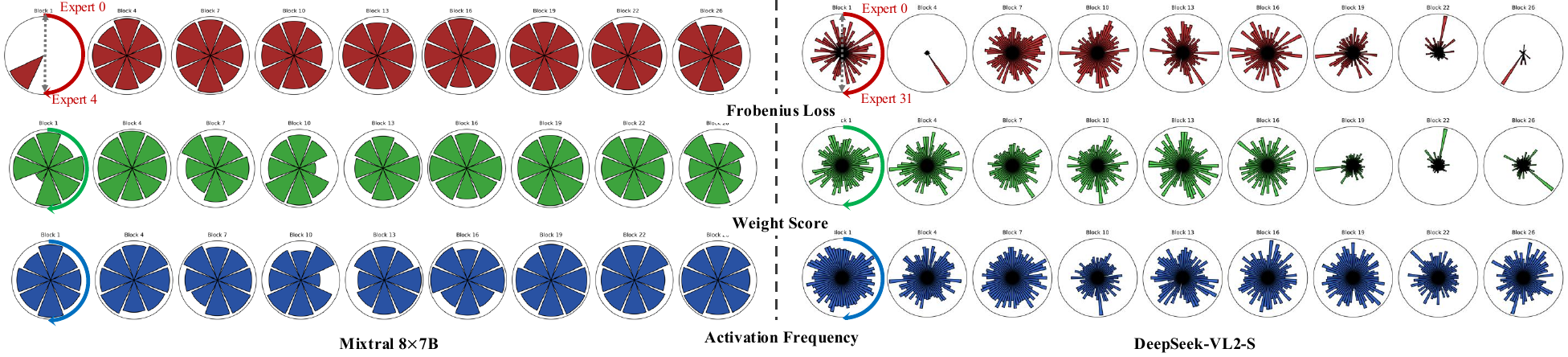}}
\caption{Comparison of experts quantization loss and activations between MoE-LLM and MoE-VLM. The left panel illustrates the quantization loss and the distribution of expert activation features for Mixtral $8\times7$b calibrated on a subset of C4 dataset~\cite{raffel2020exploring}, while the right panel presents the corresponding metrics for DeepSeek-VL2-S calibrated on a subset of the M4 dataset~\cite{li2024llava}. The expert indices are arranged in a clockwise manner, covering experts 0-8 and 0-64, respectively. Notably, the quantization loss and activation feature distributions across different experts in MoE-VLMs are significantly more imbalanced compared to those in MoE-LLMs.}
\label{fig:5}
\end{figure*}

\section{Mixture Compressor \#}
\label{sec:method}

\subsection{Preliminaries}\label{sec:pre}

\noindent\textbf{Mixture-of-Experts VLMs.} In decoder-only MoE models, conventional feed-forward networks (FFN) are replaced by MoE layer, each having $N$ experts~\cite{gale2023megablocks}. The MoE model selectively activates the $top\mbox{-}k$ experts for different tokens by a group of routing scores $\mathbf{w}_{top\mbox{-}k} = \{w_0, w_1,...,w_{k-1}\},~ k \leq N$, generated by a gating layer $\operatorname{G}(\mathbf{t})$. Fig.~\ref{fig:3}(a) also illustrates the experts selection mechanism during the inference phase based on routing scores. For instance, in the DeepSeek-VL2 model, there are 72 dynamically activated experts and two shared experts. The $i^{th}$ token in input tokens $T \in \mathbb{R}^{L\times D}$  is routed to the $top\mbox{-}6$ experts~\cite{wu2024deepseek}:
\begin{equation}\label{eq:1}
\mathbf{y} = \sum_{w_j \in Top\mbox{-}6\{\operatorname{G}(\mathbf{t_i})\}} w_{j} \operatorname{F}_j(\mathbf{t_i}) + \operatorname{F}_s(\mathbf{t_i}),
\end{equation}
where $\operatorname{F}_j$ represents the feed-forward operator of the $j\mbox{-}th$ selected expert, $\operatorname{F}_s$ represents the feed-forward operator of the shared experts and $w_j$ denotes the $j\mbox{-}th$ routing weights calculated by the gating $\operatorname{G}(\mathbf{t_i})$. Therefore, according to the definition in Eq.~\ref{eq:1}, the routing mechanism establishes the correspondence between tokens and experts. 

% Therefore, based on the definition provided in Eq~(\ref{eq:1}), selective computing is the key between input tokens and experts.
% Since the substantial memory overhead of MoE models mainly arises from the weights of its experts, quantization is employed for the experts.
\noindent \textbf{Quantization Technique.} Quantization is regarded as an effective method to compress the model weights. Specifically, floating-point weights distributed in the interval $[\mathbf{W}_{min}, \mathbf{W}_{max}]$ are mapped to the integer range of $[0,1...,2^b]$, where $b$ represents the target bit-width, and the quantization reconstruction for the weights $\mathbf{W} \in R^{n\times m}$ can be defined as:
\begin{equation}\label{eq:2}
    \mathop{\arg\min}_{\ {\mathbf{W}^q}} \ \ \| \mathbf{W}\mathbf{X}-\operatorname{Q}(\mathbf{W})\mathbf{X} \|_2^2,
\end{equation}
where $\operatorname{Q}(\cdot)$ denotes the quantization function and $\|\cdot\|_2$ is the mean square error (MSE) loss. $\operatorname{Q}(\cdot)$ is generally designed as:
\begin{equation}\label{eq:quantization}
 \left\{
    \begin{array}{lr}
     \hat{\mathbf{W}}_q = \operatorname{clamp}(\lfloor \dfrac{\mathbf{W}}{s} \rceil + z, 0, 2^N - 1),\\
     s = \dfrac{\mathbf{W}_\mathrm{max} - \mathbf{W}_\mathrm{min}}{2^b - 1}, z = - \lfloor \dfrac{\mathbf{W}_\mathrm{min}}{s} \rceil
     \end{array}
  \right.
 \end{equation}
where $\hat{\mathbf{W}}_q$ indicates quantized weight which is integer, $\lfloor \cdot \rceil$ is round operation and $\operatorname{clamp}(\cdot)$ constrains the value within integer range (e.g. $[0,1,2,3]$, $b=2$). $\Delta$ is scale factor and $z$ is quantization zero point, respectively. In 1-bit condition, weights are further quantized with binary values ($-1$ or $+1$):
\begin{equation} \label{eq:binarization}
    \begin{array}{lr}
    \hat{\mathbf{W}}_b = \operatorname{sign}(\mathbf{W}), \alpha = \frac{||\mathbf{W}||_{\ell1}}{m}\\
    \operatorname{sign}(\mathbf{W}) = 
    \begin{cases}
    1& \text{if $w \geq 0, w \in \mathbf{W}$},\\
    -1& \text{others}.
    \end{cases}
    \end{array}
\end{equation}
where $\hat{\mathbf{W}}_b$ is binary result. $\alpha$ denotes binarization scales~\cite{huang2024billm} in channel-wise manner~\cite{rastegari2016xnor}. The primary objective of this study is to explore the optimal mixture compression strategy for MoE-VLMs. To this end, we employ the efficient PTQ scheme, GPTQ~\cite{frantar2022gptq}, as our foundational tool. By utilizing Hessian-based estimation ($\mathbf{H} = 2\mathbf{X}\mathbf{X}^\top$) and quantization error compensation, GPTQ effectively reduces the group-wise quantization error of weights, enabling the quantization of DeepSeek-VL2-L within 30 minutes. Our static quantization method is orthogonal to other quantization techniques. Current PTQ methods~\cite{shao2023omniquant,lin2024awq}, codebook-based works~\cite{egiazarian2024extreme,tseng2024quip}, and even the deployment of fine-tuning~\cite{liao2024apiq} or QAT~\cite{chen2024efficientqat,liu2023llm} can be deployed for \textbf{MC\#}.

\noindent\textbf{Rule-based Experts Pruning.} 
To effectively perform dynamic experts pruning, an intuitive and efficient method involves utilizing the $top\mbox{-}k$ experts' routing scores during inference~\cite{lu2024not,huang2023towards}. This approach directly skips experts with lower routing weights among the selected set for each token. For simplicity, when $k = 2$ (as in Mixtral $8\times7$b), the pruning process follows:
\begin{equation}\label{eq:5}
\{w_0 = 0, w_1 = 1,  w_0, w_1 \in Top\mbox{-}2\{\operatorname{G}(\mathbf{t})\} \mid \frac{w_1}{w_0} < \mu \},
\end{equation}
where, $w_0$ and $w_1$ denote the gating weights of $top\mbox{-}2$ experts, respectively, with $\mu$ erving as a hyperparameter threshold for each MoE layer. This threshold is set at the median value of \( \frac{w_1}{w_0} \) derived from calibration data~\cite{lu2024not}. According to Eq.~\ref{eq:5}, when a selected expert has a significantly lower weight, it can be pruned from the current set of candidates, leaving only the primary experts for computation. However, in models like DeepSeek-VL2, which feature a larger number of experts and a greater value of $k$, rule-based approaches struggle to handle the diversity of combinations effectively.

\subsection{Pre-Loading Mixed-precision Quantization}
\label{sec:pmq}

As discussed in Sec.\ref{sec:pre}, the primary storage overhead in MoE-VLMs lies within the experts, necessitating effective compression before deployment on devices. While mainstream pruning methods often suffer from significant performance degradation under extreme pruning conditions (e.g., $\geq$50\%)\cite{frantar2023sparsegpt,sun2023simple,lu2024not}, quantization has proven to achieve substantial compression with comparatively lower performance loss~\cite{huang2024good,zhou2024survey}. However, as highlighted in~\cite{li2024examining} and our findings in Sec.~\ref{sec:3.1}, uniform bit-width quantization fails to meet the stringent accuracy requirements under extreme compression scenarios for MoE-VLMs. This limitation, coupled with the diverse and uneven characteristics of experts, motivates us to investigate optimal mixed-precision quantization strategies.

This section presents our \emph{Pre-Loading Mixed-Precision Quantization~(PMQ)} method, which aims to efficiently minimize model size through targeted expert quantization. The central objective of PMQ is to optimize the bit-width allocation strategy for experts. To achieve this, we first perform a comprehensive analysis of expert behavior on the calibration dataset, using these insights to construct an Integer Programming (IP) model that determines the optimal quantization configuration. Meanwhile, for other model components, such as attention parameters, a uniform bit-width is applied across the board.

\subsubsection{Experts Significance Analysis}\label{sec:2.2.1}

The core principle of our expert quantization strategy is to allocate bit-widths based on the relative importance of each expert within a MoE block. Initially, we analyzed the performance of various experts in the Mixtral $8\times7$b configuration, including their reconstruction loss (Frobenius norm)~\cite{he2017channel}, as well as activation patterns on the C4 dataset\cite{raffel2020exploring} and the domain-specific Math dataset~\cite{hendrycks2021measuring}. 
As illustrated in Fig.~\ref{fig:4}, the impact of experts on the model varies significantly: 1) certain experts, such as the one located at position $[2,4]$ (Fig.~\ref{fig:4}, left), exhibit minimal influence on the output activation reconstruction loss, while others in the second layer demonstrate substantially higher loss values, highlighting the imbalance among experts; 2) activation scores and frequencies follow distinct patterns, where experts at positions $[11,3]$ and $[12,7]$ show extremely low activation frequencies and average scores, while the expert at $[1,3]$ has relatively low scores but notably higher activation frequency; 3) in task-specific scenarios, such as mathematical reasoning, fewer experts are activated, resulting in a sparser distribution compared to general tasks. This variability in routing characteristics necessitates considering multiple factors when determining the optimal bit-width allocation for experts. As shown in Fig.~\ref{fig:5} (b), we further visualized the expert quantization loss and activation patterns for MoE-VLMs (e.g., DeepSeek-VL2-S) on the M4 dataset~\cite{li2024llava}, where similar trends were observed. Moreover, compared to MoE-LLMs (Fig.~\ref{fig:5} (a)), the imbalance is even more pronounced, suggesting that MoE-VLMs are better suited for mixed-precision quantization schemes.

% \begin{figure}[!t]
% % \vspace{-0.2in}
% \centerline{\includegraphics[width=0.4\textwidth]{imgs/fig_mixbit.pdf}}
% \caption{Visualization on different bit-width allocation of Mixtral 8$\times$7b (MoE-LLMs). Average bit-width is 2-bit in 32 MoE layers, each with 8 experts. Color refers to the bit size.}
% \label{fig:bit_mixtal}
% \end{figure}

% \begin{figure}
% % \vspace{-0.2in}
% \centerline{\includegraphics[width=0.4\textwidth]{imgs/fig_dpskbit.pdf}}
% \caption{Visualization on different bit-width allocation of DeepSeek-VL2-S (MoE-VLMs). Average bit-width is 2-bit in 26 MoE layers, each with 64 experts. Color refers to the bit size of each experts.}
% \label{fig:bit_dpsk}
% \end{figure}
\begin{figure*}[!t]
    \begin{minipage}[b]{.47\linewidth}
    \centering 
    \includegraphics[width=.98\columnwidth]{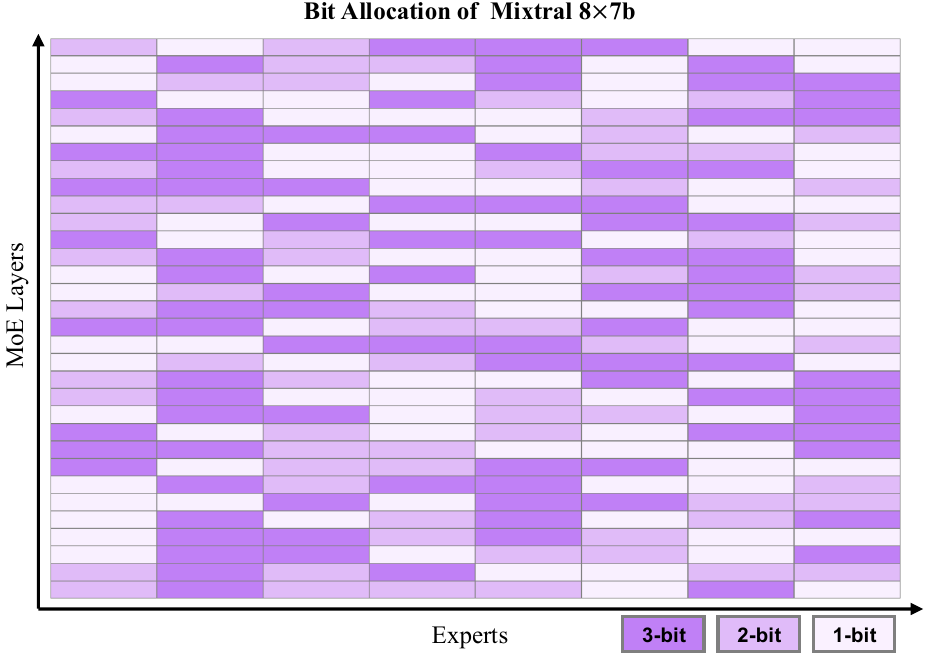}
    \caption{Visualization on different bit-width allocation of Mixtral 8$\times$7b (MoE-LLMs). Average bit-width is 2-bit in 32 MoE layers, each with 8 experts. Color refers to the bit size.}
    \label{fig:bit_mixtal}
    \end{minipage}
    \hspace{0.02\textwidth}
    \begin{minipage}[b]{.47\linewidth}
    \centering 
    \includegraphics[width=.98\columnwidth]{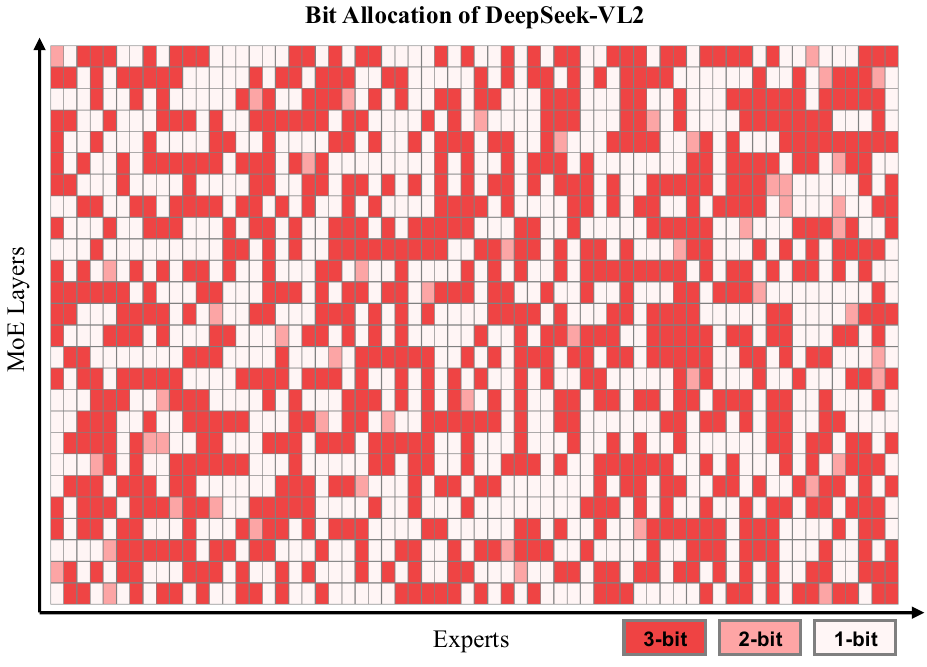}
    \caption{Visualization on different bit-width allocation of DeepSeek-VL2-S (MoE-VLMs). Average bit-width is 2-bit in 26 MoE layers, each with 64 experts. Color refers to the bit size of each expert.}
    \label{fig:bit_dpsk}
    \end{minipage}
\end{figure*}

\subsubsection{Weighted Importance Factors}\label{sec:2.2.2}

In assessing the significance of each expert in MoE-LLMs, we primarily consider two factors: access frequency and activation weight. Using an $N$-sized calibration dataset, such as C4 (general language understanding) for MoE-LLMs or M4 (visual question answering) for MoE-VLMs, inference is performed on the original 16-bit MoE models. Access frequency, defined as the rate at which an expert is activated, is expressed as $\phi_i = \frac{n_i}{N}$, where $n_i$ represents the total activations of the $i$-th expert. A higher frequency suggests greater generality and applicability across diverse tokens. However, relying solely on frequency may overlook the importance of less frequently activated experts. To address this limitation, an activation-weighted metric is introduced, calculated as $w_i = \frac{\sum_{j=1}^{N}{\sigma_j}}{N}$, where $\sigma_j$ denotes the routing weight assigned to the expert during the $j$-th inference. This metric provides a more nuanced understanding of each expert's contribution by incorporating the significance of routing weights. The overall importance of an expert is then computed as $\phi_i^{\alpha} \cdot w_i^{\beta}$, where $\alpha$ and $\beta$ are hyperparameters that balance these two parts.

% We mainly measure the significance of each expert through two factors: access frequency and activation weight. 
% Given an $N$-sized calibration dataset C4 (general language understanding dataset) for MoE-LLMs and M4 (general visual question and answer dataset), we first perform inference on the original 16-bit MoE models. For each expert, access frequency refers to the rate at which the expert is activated. Thus, $i$-th expert's access frequency is $\phi_i = \frac{n_i}{N}$, where $n_i$ is this expert's total activated number. A higher activation frequency indicates that the expert is triggered more often, suggesting its generality and applicability across a wide range of tokens. However, access frequency alone overlooks the potential significance of experts who are rarely activated.
% To account for this, we introduce the activation-weighted metric, which sums the routing weights assigned to each expert during inference. This metric for $i$-th expert can be denoted as $w_i = \frac{\sum_{j=1}^{N}{\sigma_j}}{N}$, where $\sigma_j$ is the expert's routing weight in the $j$-th inference. This provides a finer-grained measure of an expert’s contribution in MoE large models, capturing its relative importance beyond mere frequencies. The final expert significance is computed as $\phi_i^{\alpha} \cdot w_i^{\beta}$, where $\alpha$ and $\beta$ are hyperparameters used to balance the two factors.

\subsubsection{Optimal Experts Bit-Width Allocation}
Building on the determination of expert significance, we explore its application in mixed-precision quantization by assigning varying bit-widths to experts based on their importance. This approach ensures that more significant experts retain higher bit-widths, while less significant experts undergo more aggressive quantization. Beyond leveraging expert significance, we also assess the reconstruction error of output activations in each MoE layer after quantization, enabling a precise evaluation of the impact of quantizing individual experts. Specifically, for each expert, we compute the Frobenius norm (F-norm) between the model's output when the expert is quantized and the output generated when no experts are quantized, providing a quantitative measure of the effect of quantization on model performance: 
\begin{equation}
\epsilon_{i,j} = \| \mathscr{F}(\theta) - \mathscr{F}(\theta[e_i \rightarrow Q(e_i, j)]) \|_F,
\end{equation}
where $\mathscr{F}(\theta)$ is the model output with full parameters $\theta$, and $\mathscr{F}(\theta[e_i \rightarrow Q(e_i, j)])$ represents the output when only expert $e_i$ is quantized to $j$ bits. $Q(\cdot )$ denotes the quantization function.

To achieve an extremely low average bit-width of $b$ across all experts in an MoE block, with bit-width options constrained to ${1, 2, 3}$ bits, we formulate the task as an Integer Programming (IP) optimization problem. This approach allows us to efficiently compute the optimal bit-width allocation for each expert, ensuring that the targeted average bit-width is met. Remarkably, the IP optimization process is computationally efficient, completing the allocation computation within a single second. The IP model is defined as follows:

\begin{small}

\begin{equation}\label{eq:4}
\begin{aligned}
\textsc{Minimize} \quad & \sum_{i=1}^{n} \sum_{j=1}^{3} \phi_i^{\alpha} \cdot w_i^{\beta} \cdot (\epsilon_{i,j} \cdot x_{ij})^{\gamma} \\
\textsc{Subject to} \quad & \sum_{i=1}^{n} \sum_{j=1}^{3} j \cdot x_{ij} = n\cdot k, \quad \sum_{j=1}^{3} x_{ij} = 1, \quad \forall i, \\
& \sum_{i=1}^{n} x_{i3} \geq 1, \quad \sum_{i=1}^{n} x_{i2} \geq 1, \quad x_{ij} \in \{0, 1\}, \quad \forall i, j.
\end{aligned}
\end{equation}

\end{small}

In this framework, $x_{ij}$ is defined as a binary variable indicating whether the $i$-th expert is quantized to $j$ bits ($x_{ij} = 1$ if true, otherwise $x_{ij} = 0$). To ensure the accuracy of critical experts, we impose a constraint that requires at least one expert to be quantized to $3$ bits and another to $2$ bits. The parameter $\gamma$ serves as a weighting hyperparameter to balance the optimization process. Once the optimal bit-width combination for each MoE block expert is determined, the GPTQ quantization algorithm is applied to quantize the experts accordingly. For the remaining weights, such as those in the attention mechanism, gating module, and shared experts, their minimal parameter size and contribution to the output allow us to follow the recommendation from ~\cite{li2024examining} and quantize them uniformly to 4 bits, contributing an additional average bit-width of no more than 0.08 bits. Fig.~\ref{fig:bit_mixtal} and Fig.~\ref{fig:bit_dpsk} display the precision distributions of different experts.

\begin{figure*}[!t]
% \vspace{-0.2in}
\centerline{\includegraphics[width=1\textwidth]{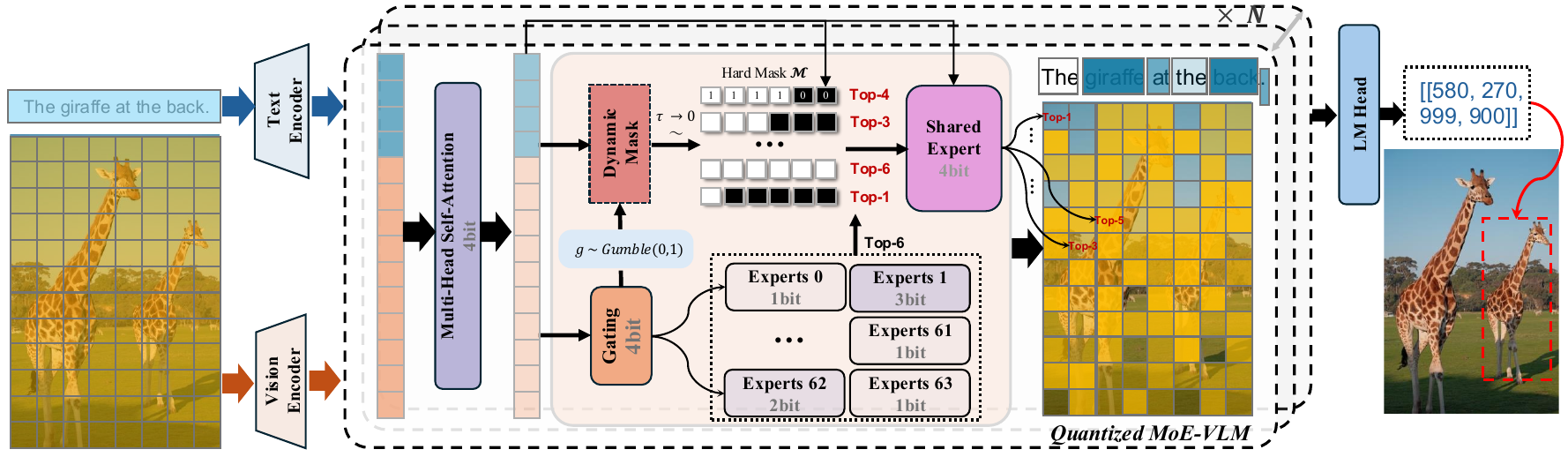}}

\caption{The workflow of Online Top-any Pruning (OTP). In this framework, each token from multimodal data is input into the quantized DeepSeek-VL2 model, where six experts are selectively activated. A learnable dynamic mask is then applied to dynamically prune redundant experts among the six, retaining only the most relevant ones for computation alongside shared experts. A key feature of OTP lies in its ability to dynamically assess the importance of tokens and flexibly prune experts, ensuring efficient and adaptive processing.}
% \vspace{-0.25in}
\label{fig:6}
\end{figure*}

\subsection{Binary Weight Saving and Dequantization}\label{1bit_pack}
This paper presents \textbf{MC\#}, which explores static compression strategies and dynamic pruning methods for MoE-LLMs in the ultra-low bit-width range, with selected static bit-width of 1-bit, 2-bit, and 3-bit. We observe that both 2-bit and 3-bit can be addressed using conventional linear quantizers, a method commonly utilized in most studies~\cite{frantar2022gptq,shao2023omniquant,huang2024slim,lin2024awq}.  We utilize the
HQQ~\cite{badri2023hqq} tool to save quantized weights and handle dequantization. In contrast, the quantization of 1-bit weights involves totally different calculations; we first provide the binarization formula for the weights:
where $\mathbf{W} \in \mathbb{R}^{d\times m}$ is the full precision weight and $\mathbf{B} \in \{-1, +1\}^{d\times m}$ denotes the binarized matrix. Due to the elements' range of $\mathbf{B}$ being ±1, we can not directly save the one-bit value into compact memory. Hence, we propose a simple transformation for $\mathbf{B}$:
\begin{equation}
    \widetilde{\mathbf{B}} = \frac{\operatorname{sign}(\mathbf{W}) + 1}{2}
\end{equation}
where $\widetilde{\mathbf{B}} \in \{0, 1\}^{d\times m}$. In this case, we can use a 1-bit memory to store each element. During the inference stage, we need to dequantize the binary weight and perform the matrix multiplication of each input vector as follows:
\begin{equation}\label{eq:10}
    s \cdot \mathbf{x}\mathbf{B} = s (\sum_{j:\widetilde{\mathbf{B}}_{ij}=1}^{d} \mathbf{x}_j  - \sum_{j:\widetilde{\mathbf{B}}_{ij}=0}^{d} \mathbf{x}_j), \operatorname{for}~i =1,2,...m
\end{equation}
where $\mathbf{x} \in \mathbb{R}^{1\times d}$ denotes one set of input token, and $s$ represents the scaling factor of each binary matrix, which is calculated from $s = \frac{\| \mathbf{W} \|_{\ell _1}}{d \times m}$~\cite{rastegari2016xnor}. In this binarized weight format, we can achieve computation without minimal multiplication. As shown in Eq.~(\ref{eq:10}), the original computation requires $dm$ multiplications and $(d-1)m$ additions, resulting in a Multiply-Accumulate Operations (MACs) consumption of $dm$ and a computational complexity of $O(m^2)$. In contrast, binary matrix operations require only $m$ multiplications and $(d-1)m$ additions, leading to a MACs consumption of just $m$ and a computational complexity of $O(m)$.

\subsection{Online Top-any Pruning}\label{sec:odr}
To address the challenges of rule-based expert pruning, we propose and introduce a learnable dynamic expert selection framework, \emph{Online Top-any Pruning (OTP)}, which can be applied to both MoE-LLMs and MoE-VLMs with any-experts. This framework further enhances inference efficiency on lightweight models. We define the pruning of $top\mbox{-}k$ selected experts as a mask selection problem, with the candidate set size denoted as $|C|$. Inspired by rule-based exploration, we prioritize masking gating weights with lower values, thus setting the size of $|C|$ to $|C|$. Given a candidate group of $|C|$ expert weights, denoted as $E \in \mathbb{R}^{1 \times k}$, the objective of dynamic pruning is to identify the optimal binary mask $\quad M^* \in \mathbb{B}^{1 \times k}$. Specifically, in the case of DeepSeek-VL2 with $k=6$, the discrete candidate set $C$ is defined as:
\begin{align}\label{eq:mask_set}
C_k &= \{M \in \mathbb{B}^{1 \times k} \ | \ 1 \leq \sum M \leq 6\} \notag \\
  &= \{\hat{M}_0, \hat{M}_1, \hat{M}_2, \hat{M}_3, \hat{M}_4, \hat{M}_5\} \notag \\
  &=\{ [1,1,1,1,1,1], [1,1,1,1,1,0], [1,1,1,1,0,0], \notag \\
  &[1,1,1,0,0,0], [1,1,0,0,0,0], [1,0,0,0,0,0]   
  \},
\end{align}
In MoE-VLMs, we first rank the $top\mbox{-}k$ experts based on their weights, assigning a value of 0 to denote the pruned experts, as shown in Fig.~\ref{fig:6}. To accurately prune unnecessary tokens while maintaining the model's performance, we define the objective function for the learnable mask in each MoE layer as follows:
\begin{equation}\label{eq:9}
\operatorname*{arg\,min}_{M_{i,j} \in C_k}  \mathcal{L}_{D}(\mathbf{X}; \{\operatorname{G}(\mathbf{t_i})_k \odot M_{i,j} \}) ,
\end{equation}
where $\operatorname{G}(\mathbf{t_{i,j}})_k$ is the $top\mbox{-}k$ index activated by original gating, $\mathcal{L}_{D}$ represents the distillation loss associated with the final output logits with non-masked MoE models, and the operator $\odot$ denotes element-wise multiplication used to prune the experts. $M_{i,j}$ is defined as the mask results of the $i^{th}$ token in the $j^{th}$ MoE layer. However, due to the non-differentiable nature of mask selection, we proceed by reformulating the mask selection process into a sampling mechanism. This transformation enables the model to learn the relationship between input tokens and the mask through optimization, making the process trainable and more adaptable.

\subsubsection{Differentiable Dynamic Experts Pruning}
To effectively model the mask sampling operation, we employ the Gumbel-Softmax~\cite{jang2016categorical,fang2024maskllm} technique to approximate the target mask matrix $M_i^*$. This approach leverages reparameterization to decompose the stochasticity of sampling into a noise variable. Specifically, it introduces a method for drawing samples from a categorical distribution $p$. The technique generates a one-hot index $y$ for sampling, thus enabling a differentiable approximation of discrete sampling operations.
\begin{equation}\label{eq:10}
    y = \operatorname{onehot}(\operatorname*{arg\,max} [log(p_i) - log(-log{\delta}_i)]),{\delta}_i \sim U(0, 1),
\end{equation}
where $- log(-log({\delta}_i))$ refers the Gumbel noise and ${\delta}_i$ is a random noise in uniform distribution. We then design a learnable router, denoted as $DM(\cdot)$ (only two linear layers of each MoE block, as shown in Tab.~\ref{tab:router}) to generate the categorical distribution $p$. To solve the issue of differentiable sampling in $\operatorname{onehot}$ and $\operatorname*{arg\,max}$ operation, the approximation of the index is further designed as:
\begin{equation}
    \hat{y}_i = \frac{\operatorname{exp} ((\operatorname{DM}(\mathbf{t_i}, w) - log(-log{\delta}_i))/ \tau)}{\sum_j \operatorname{exp} (\operatorname{DM}(\mathbf{t_i}, w) - log(-log{\delta}_i))/ \tau)}
\end{equation}
where the input of the learnable router is the token $t_i$ and its original gating weights $w$. $\tau$ refers to the temperature parameters, adjusting the sharpness of sampled results. As $t \to 0$, the predicted value $\hat{y}_i$ will asymptotically approach the one-hot format described in Eq.~\ref{eq:10}. Then, we can get $M_{i} = \hat{y}_i \times C_k$.  This operation generates a candidate expert mask based on soft indexing. As illustrated in Fig.~\ref{fig:3}(b), all operations are differentiable, making it possible to efficiently and elegantly learn an expert mask matrix by introducing a learnable router with a negligible number of additional parameters for each token in different layers.

\begin{table}
\caption{Parameter configuration of learnable router.}
	\centering
    \small
         \setlength{\tabcolsep}{3.2mm}{
     \begin{tabular}{lcrcc}\\\toprule 
        &\textbf{Model} & \textbf{FC1} & \textbf{FC2} & \textbf{Mask}\\
        \hline%第二道横线 
        \multirow{2}{*}{LLMs}&Mixtral $8\times7$b& 4096$\times$2 & 4$\times$2 & 2$\times$2\\
        &Mixtral $8\times22$b& 6144$\times$2 & 4$\times$2 & 2$\times$2\\

        \cdashline{1-5}

        \multirow{3}{*}{VLMs}&DeepSeek-VL2-L& 2569$\times$6 & 12$\times$6 & 6$\times$6\\
        &DeepSeek-VL2-S& 2048$\times$6 & 12$\times$6 & 6$\times$6\\
        &DeepSeek-VL2-T& 1792$\times$6 & 12$\times$6 & 6$\times$6\\
  
        \hline%第二道横线 
        \hline%第一道横线
        \end{tabular}}
        \label{tab:router}
\end{table}

\subsubsection{Learning Objective for MoE Model}

During the training process of the learnable mask router, we take into account both the distribution of input tokens and the expert weights derived from the original gating equation, thereby modeling a token-aware dynamic expert pruning mechanism. This approach allows the optimal distribution to be learned in an end-to-end manner. However, when solely optimizing the loss defined in Eq.~\ref{eq:9}, the model tends to converge towards learning an all-ones matrix. While the loss term $\mathcal{L}_{D}$ steadily decreases, the model exhibits a tendency to avoid masking any experts. To encourage the model to learn an appropriate dynamic expert pruning strategy under sparsity constraints, we introduce an additional sparsity regularization term:
\begin{equation} \label{eq:14}
    \mathcal{L} = \mathcal{L}_{D}(\mathbf{X}; \{\operatorname{G}(\mathbf{t_i})_k \odot M_{i,j} \}) + \lambda \|\{M\}\|_1
\end{equation}
The sparsity constraint is applied to the $\ell_1$ norm of the mask learned within the training batch, where $\lambda$ serves as a hyperparameter to balance the influence of the sparsity constraint.

\section{Experiments}
In this section, a series of experiments are conducted to evaluate the proposed \textbf{MC\#}. We present by describing the experimental parameter settings and results. In Sec.~\ref{sec:3.1}, we assess the parameter settings for the PMQ method and the performance of mixture quantization. We conduct a detailed evaluation of the performance loss and compression efficiency of OTP stage, shown in Sec.~\ref{sec:3.2}. Finally, we present the combined performance of MoE mixture compressor.

\begin{figure*}[!t]
    \begin{minipage}[b]{.47\linewidth}
    \centering 
    \includegraphics[width=.98\columnwidth]{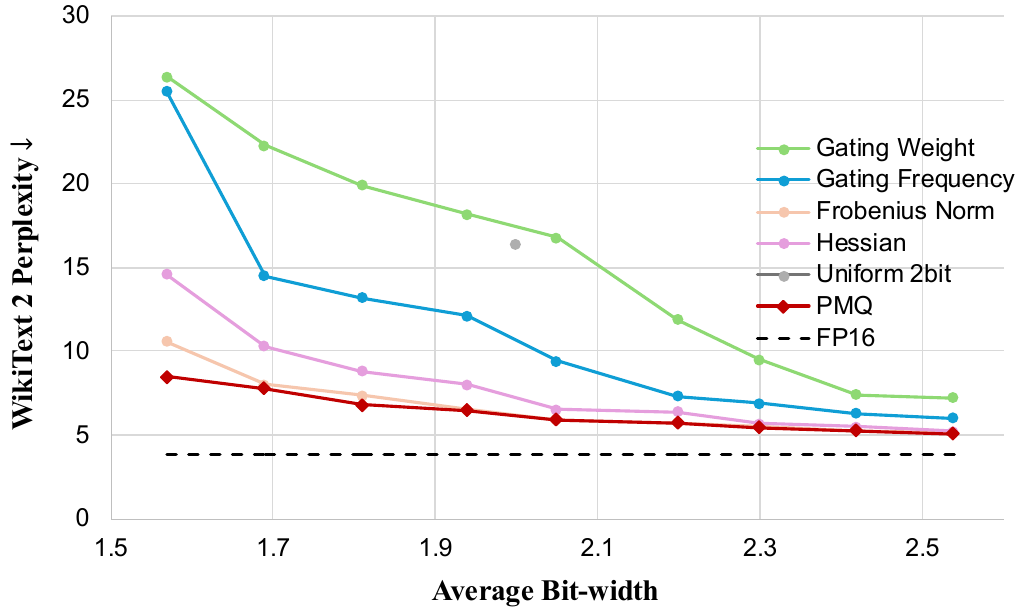}
    \caption{Quantized performance of Mixtral $8\times7$b under different mixed-precision strategies (WikiText 2 $\downarrow$).}
    \label{fig:9}
    \end{minipage}
    \hspace{0.02\textwidth}
    \begin{minipage}[b]{.47\linewidth}
    \centering 
    \includegraphics[width=.98\columnwidth]{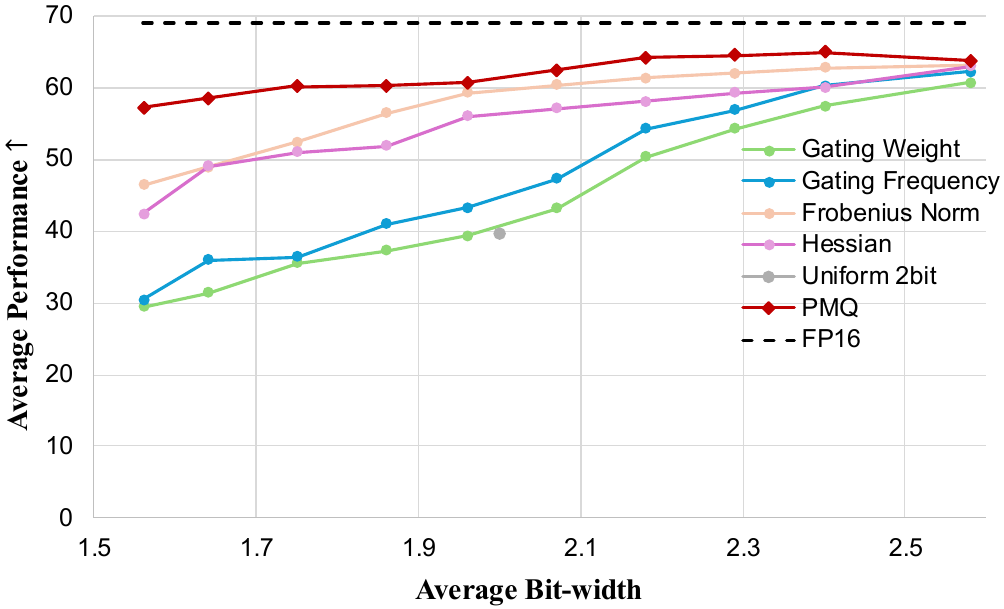}
    \caption{Quantized performance of DeepSeek-VL2-S under different mixed-precision strategies (MMBench, MMStar, MMMU, AI2D, OCRBench $\uparrow$).}
    \label{fig:10}
    \end{minipage}
\end{figure*}

\subsection{Experimental Setup}

We implemented hybrid compression deployment for both MoE-LLMs and MoE-VLMs. For the MoE-LLMs, we selected the classic open-source models Mixtral $8\times7$b and Mixtral $8\times22$b as our targets, while for the MoE-VLMs, we adopted the DeepSeek-VL2 series, including DeepSeek-VL2-L/S/T (large/small/tiny), as outlined in Tab.~\ref{tab:model}. The Mixtral $8\times7$b model can be compressed on two NVIDIA A100-80GB GPUs, whereas the Mixtral $8\times22$b model requires four NVIDIA A100-80GB GPUs. All models in the DeepSeek-VL2 series can be compressed on a single NVIDIA A100-80GB GPU. During the PMQ phase, the mixed-precision scaling factors were calibrated using the C4 dataset~\cite{raffel2020exploring}, consisting of 128 groups of randomly sampled sequences, each containing 2048 tokens. After determining the bit-width configuration, the final quantization process followed the GPTQ procedure~\cite{frantar2022gptq}. Additionally, in the dynamic pruning router learning for OTP, we trained using 4096 randomly sampled data (C4 for LLMs and M4~\cite{li2024llava} for VLMs).

% The mixed-precision factors of experts are calibrated from C4~\cite{raffel2020exploring} dataset, with 128 sets of random sequences, each 2048 tokens long. After determining the bit-width configuration, the final quantization process follows the GPTQ~\cite{frantar2022gptq} procedure. We select the open-source Mixtral $8\times7$b  and Mixtral $8\times22$b as our target models, shown in Tab.~\ref{tab:model}. Mixtral $8\times7$b can be compressed on two NVIDIA A100-80GB GPUs, while Mixtral $8\times22$b is completed on four NVIDIA A100-80GB GPUs. 

% Other None-MoE layers are set to 4-bit. Due to the significant size of expert weights, the 4-bit quantization of other parameters has minimal impact on the average bit-width. In the performance experiments for the proposed \textbf{MC}, perplexity (PPL$\downarrow$) was chosen as the metric to evaluate token prediction capabilities, primarily deploying the general text dataset WikiText2. To comprehensively assess the language capabilities of the compressed LLMs, we evaluated the models' overall abilities in eight zero-shot benchmarks ($\uparrow$) tested by EleutherAI LM Harness~\cite{gao10256836framework}.
Building on conclusions from prior studies~\cite{li2024examining}, the weights of components such as attention and shared experts were set to 4-bit precision. Since expert weights constitute the majority of parameters, quantizing other parameters to 4-bit has a negligible impact on the average bit-width, thereby maintaining an ultra-low overall bit-width. In the performance experiments conducted for the proposed \textbf{MC\#}, perplexity (PPL$\downarrow$) was chosen as the primary metric to evaluate token prediction capabilities for MoE-LLMs, using the general text dataset WikiText2. To comprehensively assess the compressed language capabilities, the overall performance of the model was further evaluated across eight zero-shot benchmarks ($\uparrow$) from the EleutherAI LM Harness~\cite{gao10256836framework}. For MoE-VLMs, six representative multimodal benchmarks ($\uparrow$) were selected, aligned with the comparison benchmarks used in DeepSeek-VL2. To ensure consistency, all VLM evaluations were conducted using the VLMEvalKit framework~\cite{duan2024vlmevalkit}.

\begin{table*}[!t]
\caption{Performance of quantized Mixtral $8\times7$b on eight zero-shot benchmarks. We deploy GPTQ as our baseline PTQ method for uniform quantization. {``}Uni" denotes the uniform quantization of 2-bit with GPTQ. Since the results of some data sets in the block score predictor (BSP)~\cite{li2024examining} method were not reported, we resumed the relevant quantized model from the official code repository and evaluated all the results under the same settings. In BSP, 25\% MoE layers are 4-bit and the left are 2-bit to achieve 2.54-bit. {``}HellaS." is the short format of {``}HellaSwag" and {``}Wino." denotes {``}Winogrande". $\downarrow$ gives the accuracy loss between quantized results and original 16-bit model.}
    \begin{center}
    \small
    \setlength{\tabcolsep}{3.2mm}
    \renewcommand{\arraystretch}{1.2}
    % \resizebox{\linewidth}{!}
    {
    \begin{tabular}{crccccccccl}
        \toprule
        \textbf{Method} & 
        Bits&
        PIQA& ARC-e& ARC-c& BoolQ& HellaS.& Wino.& MathQA & MMLU& Avg.\% $\uparrow$
        \\
        \midrule

         & 16.00&85.20& 84.01& 57.17& 85.35 & 81.48& 75.93& 39.29& 67.88& 71.29 \\
         \cline{1-11}
         Uni&3.00&82.10& 78.58& 55.80& 82.94& 79.28& 74.19& 39.26& 60.58&$69.09_{\textcolor{grey}{2.2\% \downarrow}}$ \\
         Uni&2.00&61.98& 47.20& 25.71& 62.39& 41.91 &53.22 &22.79& 30.36&$42.67_{\textcolor{grey}{28.6\% \downarrow}}$ \\
         \cdashline{1-11}
         \makecell{BSP\\ \cite{li2024examining}}&2.54&68.23&54.97&28.38&68.16&55.61&62.19&24.07&27.74&$49.07_{\textcolor{grey}{22.2\% \downarrow}}$\\
        \cdashline{1-11}
        \multirow{3}{*}{\makecell{Hessian \\ \cite{dong2020hawq}}} & 2.54&80.21&76.38&51.20&81.11&78.05&72.97&35.27&56.21&$67.18_{\textcolor{grey}{4.1\% \downarrow}}$ \\
        % &2.42&78.81&73.97&47.58&81.04&77.72&72.77&33.01&52.16&64.23  \\
        % &2.30&79.21&72.41&46.70&79.15&76.38&71.25&31.97&50.60&63.47  \\
        % &2.20&78.46&72.98&46.66&77.29&75.31&70.22&31.84&45.29&62.25  \\
        &2.05&75.32&67.26&45.01&70.29&71.90&69.11&31.07&40.85&$58.85_{\textcolor{grey}{12.4\% \downarrow}}$\\
        % &1.94&75.41&64.02&43.19&67.75&69.18&68.27&28.58&36.99&56.67  \\
        % &1.81&71.96&60.81&37.72&68.27&63.29&65.46&26.27&32.58&53.30  \\
        % &1.69&69.88&60.37&35.64&70.06&59.60&58.43&26.05&32.11&51.39  \\
        &1.57&65.26&52.12&21.84&68.21&52.91&50.32&24.99&31.58&$45.91_{\textcolor{grey}{25.4\% \downarrow}}$  \\
        \cdashline{1-11}
        \multirow{9}{*}{\textbf{PMQ}} &\cellcolor{lightmauve!40}2.54&\cellcolor{lightmauve!40}\textbf{80.52}&\cellcolor{lightmauve!40}\textbf{77.10}&\cellcolor{lightmauve!40}\textbf{51.28}&\cellcolor{lightmauve!40}\textbf{82.54}&\cellcolor{lightmauve!40}\textbf{79.03}&\cellcolor{lightmauve!40}\textbf{73.95}&\cellcolor{lightmauve!40}\textbf{39.18}&\cellcolor{lightmauve!40}\textbf{56.37}&\cellcolor{lightmauve!40}\textbf{67.50}$_{\textcolor{BrickRed}{3.8\% \downarrow}}$  \\
         & 2.42&\textbf{80.36}& \textbf{75.76}& \textbf{50.17}& \textbf{80.00}& \textbf{78.13}&\textbf{73.09}& \textbf{34.97}& \textbf{53.22}&\textbf{65.71}$_{\textcolor{grey}{5.6\% \downarrow}}$  \\
         & 2.30&\textbf{83.11}& \textbf{73.59}& \textbf{47.78}& \textbf{80.83}& \textbf{76.48}& \textbf{73.14}& \textbf{33.84}& \textbf{52.54}&\textbf{64.91}$_{\textcolor{grey}{6.4\% \downarrow}}$ \\
         & 2.20&\textbf{79.05}& \textbf{73.70}& \textbf{47.87}& \textbf{74.56}& \textbf{76.63}& \textbf{72.77}& \textbf{34.24}& \textbf{47.73}&\textbf{63.29}$_{\textcolor{grey}{8.0\% \downarrow}}$  \\
         &\cellcolor{lightmauve!40}2.05&\cellcolor{lightmauve!40}\textbf{79.16}&\cellcolor{lightmauve!40}\textbf{73.06}&\cellcolor{lightmauve!40}\textbf{48.38}&\cellcolor{lightmauve!40}\textbf{80.58}&\cellcolor{lightmauve!40}\textbf{74.95}&\cellcolor{lightmauve!40}\textbf{71.27}&\cellcolor{lightmauve!40}\textbf{31.79}&\cellcolor{lightmauve!40}\textbf{46.80}&\cellcolor{lightmauve!40}\textbf{63.25}$_{\textcolor{BrickRed}{8.0\% \downarrow}}$  \\
         & 1.94&\textbf{76.88}& \textbf{68.48}& \textbf{45.48}&\textbf{75.23}& \textbf{72.05}& \textbf{72.61}& \textbf{31.16}& \textbf{40.93}&\textbf{60.35}$_{\textcolor{grey}{10.9\% \downarrow}}$  \\
         & 1.81&\textbf{76.93}& \textbf{66.67}& \textbf{43.60}&\textbf{ 75.50}& \textbf{70.50}& \textbf{69.85}& \textbf{28.68}& \textbf{40.71}&\textbf{59.06}$_{\textcolor{grey}{12.2\% \downarrow}}$  \\
         & 1.69&\textbf{75.41}& \textbf{64.14}& \textbf{40.61}& \textbf{68.96}& \textbf{67.01}&\textbf{ 68.03}& \textbf{28.04}& \textbf{37.14}&\textbf{56.17}$_{\textcolor{grey}{15.1\% \downarrow}}$  \\
       &\cellcolor{lightmauve!40}1.57&\cellcolor{lightmauve!40}\textbf{72.42}&\cellcolor{lightmauve!40}\textbf{62.46}&\cellcolor{lightmauve!40}\textbf{37.88}&\cellcolor{lightmauve!40}\textbf{73.55}&\cellcolor{lightmauve!40}\textbf{63.17}&\cellcolor{lightmauve!40}\textbf{66.38}&\cellcolor{lightmauve!40}\textbf{26.80}&\cellcolor{lightmauve!40}\textbf{32.25}&\cellcolor{lightmauve!40}\textbf{54.49}$_{\textcolor{BrickRed}{16.8\% \downarrow}}$  \\
        \cline{1-11}
        
         % \hline
         % &16.00 & 84.93 & 83.92 & 63.40 & 87.80 & 86.07 & 79.95 & 50.12 & 74.48 & 76.33  \
          \bottomrule 
    \end{tabular}
    }
\end{center}
\label{tab:pmq_main1}
\end{table*}

% \begin{wraptable}{r}{3cm}
\begin{table}
\caption{Selected MoE-LLMs/VLMs and model configurations. Param.: the total parameter size (we set 16-bit as one parameter)(in LLM), Act Param.: activated parameter size per-token (in LLM); B: decoder block number, H: hidden dimension, E: expert number.}
	\centering
    \small
         \setlength{\tabcolsep}{1.4mm}{
     \begin{tabular}{llrcccc}\\\toprule 
        &\textbf{Model} & \textbf{Param.} & \textbf{Act Param.} & \textbf{B} & \textbf{H} & \textbf{E}\\
        \hline%第二道横线 
        \multirow{2}{*}{LLMs}&Mixtral $8\times7$b& 49b & 13b & 32 & 4096 & 8\\
        &Mixtral $8\times22$b& 141b & 39b & 56 & 6144 & 8\\

        \cdashline{1-7}

        \multirow{3}{*}{VLMs}&DeepSeek-VL2-L& 27B & 4.1B & 30 & 2569 & 72\\
        &DeepSeek-VL2-S& 16B & 2.4B & 26 & 2048 & 64\\
        &DeepSeek-VL2-T& 3B & 0.6B & 11 & 1792 & 64\\
  
        \hline%第二道横线 
        \hline%第一道横线
        \end{tabular}}
        \label{tab:model}
\end{table}
% \end{wraptable}

\subsection{Experiment on Pre-Loading Mixed-Precision Quantization}\label{sec:3.1}

% \begin{figure*}[t]
%     \begin{minipage}[b]{.49\linewidth}
%     \centering 
%     \includegraphics[width=.85\columnwidth]{imgs/fig_7.pdf}
%     \caption{Significant tokens protection of 2.05-bit Mixtral $8\times7$b. “CR” denotes the average computation compression ratio (\textcolor{RoyalBlue}{blue}); {``}PPL" denotes the perplexity(\textcolor{red}{red}); \textcolor{ForestGreen}{Star} represents the weight-only pruning performance.}
%     \label{fig:8}
%     \end{minipage}
%     \hspace{0.03\textwidth}
%     \begin{minipage}[b]{.48\linewidth}
%     \centering 
%     \includegraphics[width=.85\columnwidth]{imgs/fig_8.pdf}
%     \vspace{-0.2in}
%     \caption{Less significant tokens drop of 2.05-bit Mixtral $8\times7$b. In Mixtral $8\times7$b, we mask all experts of the less significant tokens.
%     “CR” denotes the average computation compression ratio (\textcolor{RoyalBlue}{blue}); {``}PPL" denotes the perplexity(\textcolor{red}{red}). }
%     \label{fig:9}
%     \end{minipage}
% \end{figure*}

\begin{table*}
\caption{Performance of quantized DeepSeek-VL2-L/S/T on six general multimodal benchmarks. We deploy GPTQ as our baseline PTQ method for uniform quantization. {``}Uni" denotes the uniform quantization of 2-bit with GPTQ. $\downarrow$ gives the accuracy loss between the quantized results and the original 16-bit model. The average results are calculated from MMBench, MMStar, MMMU, AI2D, and OCRBench.}
    \begin{center}
    \small
    \setlength{\tabcolsep}{2.5mm}
    \renewcommand{\arraystretch}{1.2}
    % \resizebox{\linewidth}{!}
    {
    \begin{tabular}{ccrccccccl}
        \toprule
        \textbf{Model}&\textbf{Method} & 
        Bits&
        MMBench& MMStar& MME& MMMU& AI2D& OCRBench& Avg.\% $\uparrow$
        \\
        \midrule

          \multirow{9}{*}{\textbf{DeepSeek-VL2-L}}&&16.00&82.99&62.07&1644.93&54.67&82.32&81.30&72.67\\
         \cline{2-10}
         &Uni&3.00&82.64&60.47&1637.85&48.67&82.90& 90.80&$71.01_{\textcolor{grey}{1.6\% \downarrow}}$ \\
         &Uni&2.00&60.91&49.00&1480.39&40.22&68.36 &74.50&$58.60_{\textcolor{grey}{14.1\% \downarrow}}$ \\
         \cdashline{2-10}

        &\multirow{3}{*}{\makecell{Hessian \\ \cite{dong2020hawq}}} & 2.57&80.71&60.00&1617.23&45.38&78.25&74.61&$67.79_{\textcolor{grey}{4.7\% \downarrow}}$ \\
 
        &&2.08&72.01&56.93&1578.29&42.00&75.37&69.11&$61.58_{\textcolor{grey}{11.1\% \downarrow}}$\\

        &&1.59&65.89&50.12&1580.33&40.00&73.30&63.57&$58.58_{\textcolor{grey}{14.1\% \downarrow}}$  \\
        \cdashline{2-10}
        &\multirow{3}{*}{\textbf{PMQ}} &\cellcolor{lightmauve!40}2.57&\cellcolor{lightmauve!40}\textbf{82.56}&\cellcolor{lightmauve!40}\textbf{60.00}&\cellcolor{lightmauve!40}\textbf{1628.42}&\cellcolor{lightmauve!40}\textbf{47.33}&\cellcolor{lightmauve!40}\textbf{82.29}&\cellcolor{lightmauve!40}\textbf{80.80}&\cellcolor{lightmauve!40}\textbf{70.60}$_{\textcolor{BrickRed}{2.1\% \downarrow}}$  \\
         &&\cellcolor{lightmauve!40}2.08&\cellcolor{lightmauve!40}\textbf{79.21}&\cellcolor{lightmauve!40}\textbf{56.93}&\cellcolor{lightmauve!40}\textbf{1611.77}&\cellcolor{lightmauve!40}\textbf{44.00}&\cellcolor{lightmauve!40}\textbf{81.31}&\cellcolor{lightmauve!40}\textbf{78.30}&\cellcolor{lightmauve!40}\textbf{67.95}$_{\textcolor{BrickRed}{4.7\% \downarrow}}$  \\
       &&\cellcolor{lightmauve!40}1.59&\cellcolor{lightmauve!40}\textbf{74.83}&\cellcolor{lightmauve!40}\textbf{54.30}&\cellcolor{lightmauve!40}\textbf{1512.71}&\cellcolor{lightmauve!40}\textbf{46.67}&\cellcolor{lightmauve!40}\textbf{78.17}&\cellcolor{lightmauve!40}\textbf{73.70}&\cellcolor{lightmauve!40}\textbf{65.35}$_{\textcolor{BrickRed}{7.3\% \downarrow}}$  \\
        \cline{1-10}

          \multirow{9}{*}{\textbf{DeepSeek-VL2-S}}&&16.00&78.61&56.73&1654.30&47.33&79.05&83.50&69.04\\
         \cline{2-10}
         &Uni&3.00&78.69&54.00&1669.77&42.00&77.42&81.80&$66.78_{\textcolor{grey}{2.3\% \downarrow}}$ \\
         &Uni&2.00&34.62&33.47&1247.61&23.33&48.77&57.00&$39.44_{\textcolor{grey}{29.6\% \downarrow}}$ \\
         \cdashline{2-10}

        &\multirow{3}{*}{\makecell{Hessian \\ \cite{dong2020hawq}}} & 2.58&72.88&52.27&1613.41&40.67&67.84&74.31&$61.79_{\textcolor{grey}{7.3\% \downarrow}}$ \\
 
        &&2.07&66.67&48.83&1498.27&36.77&67.53&72.52&$58.46_{\textcolor{grey}{10.6\% \downarrow}}$\\

        &&1.58&57.28&41.05&1378.27&31.62&67.77&69.24&$53.59_{\textcolor{grey}{15.5\% \downarrow}}$  \\
        \cdashline{2-10}
        &\multirow{3}{*}{\textbf{PMQ}} &\cellcolor{lightmauve!40}2.58&\cellcolor{lightmauve!40}\textbf{74.05}&\cellcolor{lightmauve!40}\textbf{54.33}&\cellcolor{lightmauve!40}\textbf{1674.45}&\cellcolor{lightmauve!40}\textbf{43.67}&\cellcolor{lightmauve!40}\textbf{69.53}&\cellcolor{lightmauve!40}\textbf{76.70}&\cellcolor{lightmauve!40}\textbf{63.66}$_{\textcolor{BrickRed}{5.4\% \downarrow}}$  \\
         &&\cellcolor{lightmauve!40}2.07&\cellcolor{lightmauve!40}\textbf{69.16}&\cellcolor{lightmauve!40}\textbf{51.27}&\cellcolor{lightmauve!40}\textbf{1576.86}&\cellcolor{lightmauve!40}\textbf{42.33}&\cellcolor{lightmauve!40}\textbf{72.53}&\cellcolor{lightmauve!40}\textbf{76.70}&\cellcolor{lightmauve!40}\textbf{62.40}$_{\textcolor{BrickRed}{6.6\% \downarrow}}$  \\
       &&\cellcolor{lightmauve!40}1.58&\cellcolor{lightmauve!40}\textbf{63.32}&\cellcolor{lightmauve!40}\textbf{48.47}&\cellcolor{lightmauve!40}\textbf{1470.23}&\cellcolor{lightmauve!40}\textbf{35.77}&\cellcolor{lightmauve!40}\textbf{67.26}&\cellcolor{lightmauve!40}\textbf{71.30}&\cellcolor{lightmauve!40}\textbf{57.23}$_{\textcolor{BrickRed}{11.8\% \downarrow}}$  \\
        \cline{1-10}

         \multirow{9}{*}{\textbf{DeepSeek-VL2-T}}&&16.00&72.08&49.47&1554.01&39.89&74.77&80.30&63.30\\
         \cline{2-10}
         &Uni&3.00&67.10&46.53&1560.91&38.33&70.85&77.4&$60.04_{\textcolor{grey}{3.3\% \downarrow}}$ \\
         &Uni&2.00&1.12&1.00&448.38&0.00&4.93&0.00&$1.41_{\textcolor{grey}{61.9\% \downarrow}}$ \\
         \cdashline{2-10}

        &\multirow{3}{*}{\makecell{Hessian \\ \cite{dong2020hawq}}} & 2.59&66.68 & 42.36 & 1566.27 & 29.79 & 69.20 & 70.02&$55.61_{\textcolor{grey}{7.7\% \downarrow}}$ \\
 
        &&2.12&54.33&39.06&1490.27&30.00&57.46&71.03&$50.58_{\textcolor{grey}{12.7\% \downarrow}}$\\

        &&1.63&25.51&30.02&998.70&15.03&30.00&66.56&$33.42_{\textcolor{grey}{29.88\% \downarrow}}$  \\
        \cdashline{2-10}
        &\multirow{3}{*}{\textbf{PMQ}} &\cellcolor{lightmauve!40}2.59&\cellcolor{lightmauve!40}\textbf{70.10}&\cellcolor{lightmauve!40}\textbf{46.93}&\cellcolor{lightmauve!40}\textbf{1564.56}&\cellcolor{lightmauve!40}\textbf{34.00}&\cellcolor{lightmauve!40}\textbf{72.15}&\cellcolor{lightmauve!40}\textbf{74.10}&\cellcolor{lightmauve!40}\textbf{59.45}$_{\textcolor{BrickRed}{3.8\% \downarrow}}$  \\
         &&\cellcolor{lightmauve!40}2.12&\cellcolor{lightmauve!40}\textbf{61.42}&\cellcolor{lightmauve!40}\textbf{39.73}&\cellcolor{lightmauve!40}\textbf{1515.26}&\cellcolor{lightmauve!40}\textbf{31.33}&\cellcolor{lightmauve!40}\textbf{63.99}&\cellcolor{lightmauve!40}\textbf{70.90}&\cellcolor{lightmauve!40}\textbf{53.47}$_{\textcolor{BrickRed}{9.8\% \downarrow}}$  \\
       &&\cellcolor{lightmauve!40}1.63&\cellcolor{lightmauve!40}\textbf{25.51}&\cellcolor{lightmauve!40}\textbf{34.67}&\cellcolor{lightmauve!40}\textbf{1230.16}&\cellcolor{lightmauve!40}\textbf{25.52}&\cellcolor{lightmauve!40}\textbf{39.44}&\cellcolor{lightmauve!40}\textbf{63.10}&\cellcolor{lightmauve!40}\textbf{37.65}$_{\textcolor{BrickRed}{25.7\% \downarrow}}$  \\
        \cline{1-10}
        
          \bottomrule 
    \end{tabular}
    }
\end{center}
\label{tab:pmq_main2}
\end{table*}

\subsubsection{Ablation of Bit-Width Allocating Metrics}

% Fig.~\ref{fig:5_1} illustrates a significant decline in model performance with random bit-width allocation. And employing only the routing scores of experts from calibration data, the curve in Fig.~\ref{fig:5_2} though better than random allocation, the PPL curve is still high. However, activation frequencies, in comparison to weight, shows a better performance. 

% Furthermore, in conventional networks and dense LLMs, Hessian-based quantization loss is a common use for bit-width allocation~\cite{dong2020hawq,huang2024slim}. We also utilized is as a compared metric for expert-wise bit-width allocation. Fig.~\ref{fig:5_2} also contains three metric curves: Hessian, F-norm, and PMQ. F-norm and PMQ are more effective than Hessian for expert-wise bit-width allocation, exhibiting better performance under different bit-widths. When the average bit-width exceeds 2-bit, the F-norm is similar to the PPL curve of PMQ; below 2-bit, the lead of PMQ gradually widens.

To demonstrate the advantages of our PMQ method, we compare the performance of the PMQ quantization curve with results obtained using only gating weights, gating frequencies, Frobenius norm, Hessian-based method, and uniform 2-bit quantization. Fig.~\ref{fig:9} and Fig.~\ref{fig:10} illustrate this comparison by presenting the PPL performance on Mixtral $8\times7$b and the average results across five general benchmarks for DeepSeek-VL2-S, respectively. While using expert routing scores derived solely from calibration data shows an improvement over random assignment, the PPL curve remains relatively high. However, activation frequency outperforms weights in achieving better performance.

Moreover, in traditional neural networks and dense LLMs, Hessian-based bit-width allocation has been a commonly effective strategy~\cite{dong2020hawq,huang2024slim}. We also adopt this as a comparative baseline for expert bit-width allocation. Fig.~\ref{fig:9} and Fig.~\ref{fig:10} include three benchmark curves: Hessian, Frobenius norm (F-norm), and PMQ. For experts bit-width allocation, both F-norm and PMQ outperform Hessian, consistently delivering better performance across various bit-widths. When the average bit-width exceeds 2-bit, the PPL curves of F-norm and PMQ appear similar; however, below 2-bit, PMQ's advantage becomes increasingly significant. This suggests that PMQ is particularly effective in identifying optimal mixed-precision strategies for MoE under ultra-low bit-width settings.

\subsubsection{Comparison of Mixed-Precision Quantization}

% We present a comprehensive comparison of the performance of \emph{PMQ} within the ultra-low bit-width range. GPTQ was set as the baseline for uniform bit-width quantization, denoted as {``}Uni" in Tab.~\ref{tab:1}. We also compare it with a recent mixed-precision approach for MoE-LLMs known as the block score predictor (BSP)~\cite{li2024examining}. Following Eq.~\ref{eq:4}, we set the average bit-width of Mixtral $8\times7$b within the range of 1.57-bit to 2.54-bit. As shown in Tab.~\ref{tab:1}, the uncompressed 16-bit model achieves an average accuracy of 71.29\%. With uniform precision quantization, the average loss for the 3-bit model is approximately 2.2\%, while the loss for the 2-bit model increases significantly by 28.6\%, highlighting the challenges in maintaining model accuracy with existing uniform precision quantization methods at ultra-low bit widths. 

We conducted a comprehensive evaluation of the performance of \emph{PMQ} in the ultra-low bit-width range. GPTQ was used as the baseline for uniform bit-width quantization, denoted as {``}Uni" in Tab.~\ref{tab:pmq_main1} and Tab.~\ref{tab:pmq_main2}. Additionally, we compared PMQ against a recent mixed-precision approach for MoE-based large language models (MoE-LLMs), known as the Block Sparsity Predictor (BSP)~\cite{li2024examining}. The experiments were performed on Mixtral $8\times7$b and the DeepSeek-VL2 models (L/S/T variants), with the average selectable expert bit-widths set between 1.5 and 2.5 bits. The final results, presented as the average bit-widths across all language backbones, are reported in the tables. As shown in Tab.~\ref{tab:pmq_main1} and Tab.~\ref{tab:pmq_main2}, for both MoE-LLMs and MoE-VLMs, uniform precision quantization results in average performance losses of 1.3–3.3\% for 3-bit models. However, for 2-bit models, the performance degradation becomes significantly more pronounced, with losses approaching 30\%. This highlights the substantial challenges of maintaining model accuracy using existing uniform precision quantization methods under ultra-low bit-width settings.

\begin{figure*}[!t]
    \begin{minipage}[b]{.47\linewidth}
    \centering 
    \includegraphics[width=.98\columnwidth]{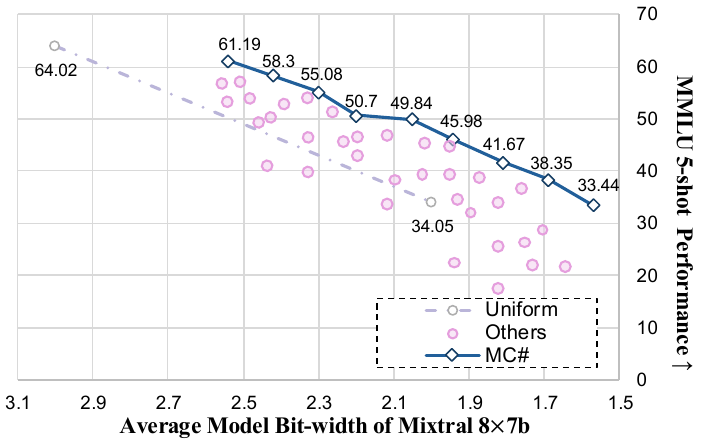}
    \caption{Pareto curve of performance-precision trade-offs on Mixtral $8\times7$b model. {``}Others" denotes the other mixed-precision method, such as random configuration and other method in Fig.~\ref{fig:9}.}
    \label{fig:mixtral_pareto}
    \end{minipage}
    \hspace{0.02\textwidth}
    \begin{minipage}[b]{.47\linewidth}
    \centering 
    \includegraphics[width=.98\columnwidth]{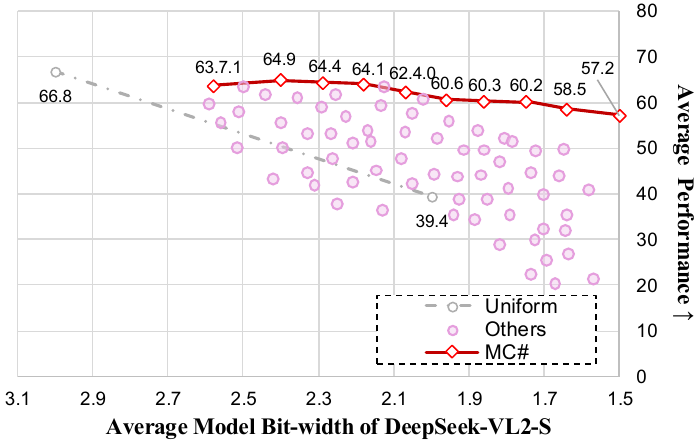}
    \caption{Pareto curve of performance-precision trade-offs on DeepSeek-VL2-S model. {``}Others" denotes the other mixed-precision method, such as random configuration and other method in Fig.~\ref{fig:10}.}
    \label{fig:dpsk_pareto}
    \end{minipage}
\end{figure*}

Under a 2.54-bit setting, the average accuracy of the BSP algorithm is only 49.07\%. In contrast, the proposed \emph{PMQ} algorithm achieves an accuracy of 67.50\%, exceeding BSP by 18.4\% and only falling off the 16-bit Mixtral $8\times7$b by 3.8\%. Notably, \emph{PMQ} can maintain an accuracy of 54.49\% at 1.57-bit, even outperforming BSP at 2.54-bit by 5.4\%. Methods based on the Hessian algorithm, as shown in Tab.~\ref{tab:pmq_main1}, consistently underperform compared to \emph{PMQ} across different bit-widths. Specifically, while the Hessian algorithm lags by a marginal 0.2\% at 2.54 bits, \emph{PMQ} demonstrates a more pronounced advantage below 2 bits, leading by 9.4\% at 1.57 bits. Further insights are provided in Tab.~\ref{tab:pmq_main2}, where the \emph{PMQ} algorithm achieves the best performance under identical low-bit settings across DeepSeek-VL2 models with varying parameter scales. For example, in the DeepSeek-VL2-L model, \emph{PMQ} achieves a score of 70.60\% at 2.57 bits, only 2.1\% lower than the fp16 baseline, while reducing the model size to just 16\% of the original. Additionally, it outperforms the Hessian-based algorithms. Notably, as the model size increases, quantization loss decreases, a trend corroborated by the curves in Fig.~\ref{fig:2}, which collectively indicate that compressed large-parameter MoE-VLMs provide greater benefits compared to small full-precision models.

\begin{table*}[t]
\caption{Ablation evaluation of \emph{PMQ} and \emph{OTP} for MMoE-LLMs and MoE-VLMs. {``}Params" denotes the parameter size, and {``}Act Params" is the averaged activated parameters for one token. The parameter calculation of the compressed model includes the compressed weights and quantizer parameters (e.g., scaling factor and zero factor for dequantization). We carry out the average activated parameter size and speedup on the C4 and M4 datasets. 16-bit Mixtral $8\times7$b uses 2 A100-80GB GPUs, and Mixtral $8\times22$b uses 4. Other quantized MoE models are tested on one A100-80GB GPU.}
    \begin{center}
    \small
    \setlength{\tabcolsep}{2.0mm}
    \renewcommand{\arraystretch}{1.2}
    % \resizebox{\linewidth}{!}
    {
    \begin{tabular}{lrcccrrrrr}
        \toprule
        \textbf{LLMs} & 
        Bits&
        \textbf{PMQ}& \textbf{OTP}& \textbf{Uni} & LM-Eval\% $\uparrow$ &VLM-Eval\% $\uparrow$ &Params.(GB) &Act Params.(GB) & Speedup 
        \\
        \midrule

         \multirow{4}{*}{Mixtral $8\times7$b}  &16.00 & - & - &-& 71.29 &-& 96.80 & 26.31 & $1.00\times$  \\
        &2.00 & - & - &\checkmark& 42.67 &-& 13.61 & 3.70 & $1.72\times$   \\
         \cline{2-10}

        & 2.05&\checkmark&-& -& 63.25&-& 13.41&3.73& $1.67\times$\\
        &\cellcolor{lightmauve!40}\textbf{2.05}&\cellcolor{lightmauve!40}\checkmark&\cellcolor{lightmauve!40}\checkmark&\cellcolor{lightmauve!40}-&\cellcolor{lightmauve!40}\textbf{62.68}&\cellcolor{lightmauve!40}-&\cellcolor{lightmauve!40}\textbf{13.41}&\cellcolor{lightmauve!40} \textbf{3.23}&\cellcolor{lightmauve!40}\textbf{1.80}$\times$\\
       
        \cline{1-10}
        \multirow{4}{*}{Mixtral $8\times22$b}&16.00 & - & - &-& 76.33 &-& 281.24 & 76.49 & $1.00\times$  \\
        &2.00 & - & - &\checkmark&50.44 &-&38.08&10.35&$1.95\times$\\
         \cline{2-10}
         
        & 2.05&\checkmark&-&-&67.94&-&38.35& 10.42&1.80$\times$\\
        &\cellcolor{lightmauve!40} \textbf{2.05}&\cellcolor{lightmauve!40}\checkmark&\cellcolor{lightmauve!40}\checkmark&\cellcolor{lightmauve!40}-&\cellcolor{lightmauve!40} \textbf{66.50}&\cellcolor{lightmauve!40}-&\cellcolor{lightmauve!40}\textbf{38.35}&\cellcolor{lightmauve!40} \textbf{9.03}&\cellcolor{lightmauve!40}\textbf{1.87}$\times$\\

         \cline{1-10}
         \multirow{4}{*}{DeepSeek-VL2-S}&16.00 & - & - &-&-&69.04&32.39&4.95& $1.00\times$  \\
        &2.00 & - & - &\checkmark&-&39.43&4.54&0.81&$1.62\times$\\
         \cline{2-10}
         
        &2.07&\checkmark&-&-&-&62.40&4.59&0.84&1.62$\times$\\
        &\cellcolor{lightmauve!40} \textbf{2.07}&\cellcolor{lightmauve!40}\checkmark&\cellcolor{lightmauve!40}\checkmark&\cellcolor{lightmauve!40}-&\cellcolor{lightmauve!40}-&\cellcolor{lightmauve!40}\textbf{60.92}&\cellcolor{lightmauve!40}\textbf{4.59}&\cellcolor{lightmauve!40} \textbf{0.52}&\cellcolor{lightmauve!40}\textbf{1.92}$\times$\\

        \cline{1-10}
         \multirow{4}{*}{DeepSeek-VL2-L}&16.00 & - & - &-&-&72.67&54.95&8.27& $1.00\times$  \\
        &2.00 & - & - &\checkmark&-&58.60&7.85&1.18&$1.78\times$\\
         \cline{2-10}
         
        &2.08&\checkmark&-&-&-&67.95&7.92&1.19&1.77$\times$\\
        &\cellcolor{lightmauve!40} \textbf{2.08}&\cellcolor{lightmauve!40}\checkmark&\cellcolor{lightmauve!40}\checkmark&\cellcolor{lightmauve!40}-&\cellcolor{lightmauve!40}-&\cellcolor{lightmauve!40}\textbf{66.42}&\cellcolor{lightmauve!40}\textbf{7.92}&\cellcolor{lightmauve!40} \textbf{0.79}&\cellcolor{lightmauve!40}\textbf{1.86}$\times$\\
    \bottomrule 
    \end{tabular}}
\end{center}
\label{tab:3}
\end{table*}

% \begin{figure}[!t]
% % \vspace{-0.2in}
% \centerline{\includegraphics[width=0.4\textwidth]{imgs/fig_pareto2.pdf}}
% \caption{Pareto curve of performance-precision trade-offs on Mixtral $8\times7$b model. {``}Others" denotes the other mixed-precision method, such as random configuration and other method in Fig.~\ref{fig:9}.}
% \label{fig:mixtral_pareto}
% \end{figure}

% \begin{figure}[!t]
% % \vspace{-0.2in}
% \centerline{\includegraphics[width=0.4\textwidth]{imgs/fig_pareto.pdf}}
% \caption{Pareto curve of performance-precision trade-offs on DeepSeek-VL2-S model. {``}Others" denotes the other mixed-precision method, such as random configuration and other method in Fig.~\ref{fig:10}.}
% \label{fig:dpsk_pareto}
% \end{figure}

\begin{figure}[!t]
% \vspace{-0.2in}
\centerline{\includegraphics[width=0.45\textwidth]{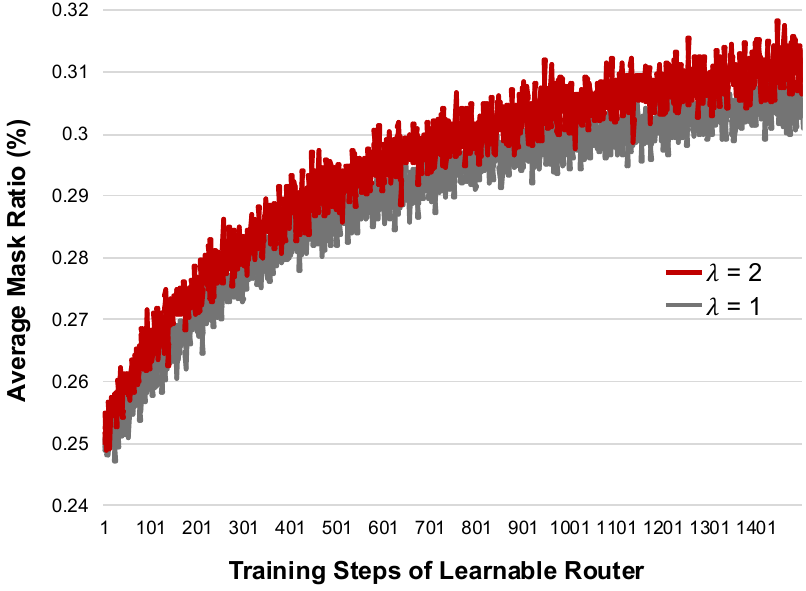}}
\caption{Ablation of mask ratio during training. We train our proposed learnable router with the DeepSeek-VLs2-S model under different $\lambda$.}
\label{fig:zero_training}
\end{figure}

\subsubsection{Pareto Advantage of PMQ}

When compressing models to an ultra-low bit-width range of 1.5–2.5 bits, we can flexibly balance the required bit-width, model size, and performance demands. Therefore, exploring the compression Pareto frontier under different bit-widths becomes particularly critical~\cite{liu2025paretoq}. Fig.~\ref{fig:mixtral_pareto} and Fig.~\ref{fig:dpsk_pareto} compare the proposed PMQ strategy with other random mixed-precision strategies across various bit-widths in the low-bit range. The results demonstrate that, for both MoE-LLMs and MoE-VLMs, PMQ consistently achieves the optimal Pareto frontier under different bit-widths. Furthermore, we observed that due to the more pronounced sparsity of MoE-VLMs (as observed in Fig.~\ref{fig:5}), the Pareto curve for DeepSeek-VL2-S is notably flatter. This indicates that MoE-VLMs are inherently more suitable for mixed-precision compression.

\begin{table}
\caption{Ablation of the combination of \textbf{MC\#}. {``}Pruning Ratio" denotes the average expert's pruning percentage during the evaluation. We evaluate the MoE-LLMs on the WikiText2 dataset with PPL$\downarrow$, MoE-VLMs on MMBench, MMStar, MMMU, AI2D, OCRBench with average score $\uparrow$.}
	\centering
    \small
         \setlength{\tabcolsep}{0.9mm}{
     \begin{tabular}{llcccc}\\\toprule 
        \textbf{Model}&\textbf{Method} & \makecell{\textbf{Pruning} \\ \textbf{Ratio(\%)}} & \textbf{Bits} & \textbf{PPL$\downarrow$} & \textbf{Score$\uparrow$} \\
        \hline%第二道横线 
        \multirow{4}{*}{Mixtral $8\times7$b}&PMQ&0&2.05&5.91&-\\
        &PMQ&0&1.69&7.78&-\\
        &PMQ+ODP~\cite{huang2024mc}&21.27&2.05&6.62&-\\
        &\textbf{PMQ+OTP}&\textbf{23.10}&\textbf{2.05}&\textbf{6.45}&-\\

        \cdashline{1-6}

        \multirow{4}{*}{DeepSeek-VL2-S}&PMQ&0&2.07&-&62.40\\
        &PMQ&0&1.64&-&58.46\\
        &PMQ+random&16.67&2.07&-&52.69\\
        &\textbf{PMQ+OTP}&\textbf{33.51}&\textbf{2.07}&-&\textbf{60.92}\\
  
        \hline%第二道横线 
        \hline%第一道横线
        \end{tabular}}
        \label{tab:ablation_mc}
\end{table}

\subsection{Experiment on Online Top-any Pruning}\label{sec:3.2}

In the pre-loading phase, \emph{PMQ} enables the compression of MoE-LLMs to an exceptionally low bit-width range. Furthermore, during the inference phase, we apply the \emph{OTP} outlined in Sec.~\ref {sec:odr} to the quantized MoE model, further enhancing the efficiency of real-time inference for lightweight models.

\subsubsection{Sparsity Constrain}
We first observed the impact of different mask sparsity parameters, denoted as $\lambda$, on the training process. As shown in Fig.~\ref{fig:zero_training}, due to the gradient direction constraints introduced in Eq.~\ref{eq:14}, the model tends to adopt a higher mask ratio while minimizing the distillation loss as much as possible. When $\lambda=1$, the model progressively increases the average mask ratio during training to approximately 30\%. As $\lambda$ increases (from 1 to 2 in Fig.~\ref{fig:zero_training}), the mask ratio also rises accordingly, which aligns with the definition in Eq.~\ref{eq:14}. All OTP results reported in this study are based on $\lambda=1$.

\subsubsection{Comparison of Online Experts Pruning Methods}
As shown in Tab.~\ref{tab:ablation_mc}, the OTP method achieves a PPL of 6.45 on a 2.05-bit model with an additional expert masking ratio of 23.10\%, whereas the rule-based ODP~\cite{huang2024mc} only reaches a PPL of 6.62 with a 21.27\% masking ratio. Moreover, as the $k$ value in $top\mbox{-}k$ increases within MoE-VLMs, rule-based methods like ODP struggle to perform effective pruning across the numerous possible expert combinations. In contrast, OTP enables dynamic expert masking through a flexible, soft, and learnable approach. For instance, OTP achieves dynamic pruning of 33.51\% of experts in the 2.07-bit DeepSeek-VL2-S model, with only a 1.5\% drop in the multimodal benchmark score. Interestingly, we observed that rather than further compressing model parameter size (from 2.07-bit to 1.64-bit), the PMQ+OTP configuration activates fewer total parameters during dynamic inference, resulting in less performance drop. 

\subsubsection{Performance on Challenging Language Benchmarks}

In this section, we expand our experiments on more challenging datasets in Tab.~\ref{tab:a_challenging}, considering the importance of performance testing on more complex long text or reasoning capabilities of compressed MoE model~\cite{cobbe2021training,bai2023longbench,chen2021evaluating}. We have observed that in challenging tasks like GSM8K, HumanEval, and long-context Needle-in-a-haystack, the performance drop of model compression becomes more pronounced. This phenomenon holds true in other MoE-LLM compression methods~\cite{frantar2022gptq,huang2024slim,shao2023omniquant,lu2024not} as well. However, our \emph{PMQ} method, compared to the latest method like BSP~\cite{li2024examining} and HAWQ~\cite{dong2020hawq} with Hessian-based approaches for MoE-LLM, is still able to maintain state-of-the-art performance.

\begin{table}[t]
% \vspace{-0.1in}
\caption{{Comparison of different mixed-precision quantization methods on challenging benchmarks. NIAH denotes the task in Needle-in-a-haystack, which is a more challenging task for evaluating long-context ability.}}
    \begin{center}
    \setlength{\tabcolsep}{2.3mm}

    \small
    % \resizebox{\linewidth}{!}
    {
    \begin{tabular}{crrrr}
        \toprule
        \textbf{Method} & 
        Bits&
        GSM8K$\uparrow$&\makecell{HumanEval \\(p@10)}$\uparrow$&NIAH$\uparrow$\\
        \midrule
        &16.00&58.30&59.15&100.00\\
        \midrule
        Uniform&3.00&38.13&29.88&98.48\\
        Uniform&2.00&0.00&0.00&0.00\\
        \midrule
        BSP&2.54&4.25&3.21&42.21\\
        Hessian&2.54&33.59&25.49&100.00\\
        Hessian&2.05&17.24&7.84&93.45\\
        \midrule
        \textbf{PMQ}&2.54&\textbf{37.67}&\textbf{29.34}&\textbf{100.00}\\
        \textbf{PMQ+OTP}&2.54&\textbf{36.44}&\textbf{27.92}&\textbf{100.00}\\
        \cdashline{1-5}
        \textbf{PMQ}&2.05&\textbf{19.97}&\textbf{11.83}&\textbf{100.00}\\
        \textbf{PMQ+OTP}&2.05&\textbf{20.91}&\textbf{11.71}&\textbf{100.00}\\
        
        \bottomrule 
    \end{tabular}
    }

\end{center}
\label{tab:a_challenging}
\end{table}

\subsection{Memory Saving and Inference Efficiency}

% Tab.~\ref{tab:3} details the memory compression, speed tests, and average results(LM-Eval and VLMEval) of the proposed \textbf{MC\#}. We utilize the HQQ~\cite{badri2023hqq} tool to save quantized weights and handle dequantization. To saving the binary weight, we design a bit-change transformation for this 1-bit format. After applying PMQ, the Mixtral $8\times7$b model can be compressed to a memory from 10.82 to 6.65 GB. During dynamic inference, ODP reduces activation parameters by about 15\%, with average accuracy decreasing by less than 1\%. At 2.05-bit, the average activation parameter per token is only 3.23 GB, resulting in a $1.80\times$ increase in inference speed and an evaluation accuracy of 62.68\%. Tab.~\ref{tab:3} also compares the performance of the LLaMA series dense models. The MC compressed 2.54-bit Mixtral $8\times7$b model outperforms the 26.03 GB 16-bit LLaMA2-13b model, with a total parameter size of 16.65 GB and activation parameters of 3.69 GB. We have also extended the compression experiments to the Mixtral $8\times22$b model. MC shows higher overall performance compared to mainstream dense models, without any training of original model.

Tab.~\ref{tab:3} provides a detailed overview of the proposed \textbf{MC\#} framework, including memory compression, speed benchmarks, and average performance results (through LM-Eval and VLMEval). We utilized the HQQ~\cite{badri2023hqq} tool for quantized weight storage and dequantization, and specifically designed a 1-bit transformation format for storing binary weights. For MoE-LLMs, we sampled 128 data points with a token length of 2048 from the C4 dataset. For MoE-VLMs, we randomly sampled 128 VQA data from the M4 dataset. The speedup was ultimately calculated based on the average token generation time. Following the application of PMQ’s 2.05-bit quantization, the memory usage of the Mixtral $8\times7$b model was reduced from 96.8GB to 13.41GB, while DeepSeek-VL2-S was compressed from 32.39GB to 4.59GB. During dynamic inference, OTP further reduced activated parameters by approximately 20–33\%, with an average accuracy loss of only about 1\%. Additionally, PMQ, leveraging the HQQ-based quantization framework and ATEN kernels, achieved a 1.6–2x speedup, while OTP provided an additional 10–20\% improvement in overall inference speed. Similar trends were observed across MoE-LLMs and MoE-VLMs of varying scales, further highlighting the effectiveness and scalability of the proposed approach.

\begin{table}[t]
\vspace{-0.1in}
\caption{{Latency comparison of MoE and dense LLM under different hardware platforms.}}
    \begin{center}
    \setlength{\tabcolsep}{2.0mm}
    \small

    % \resizebox{\linewidth}{!}
    {
    \begin{tabular}{crrrr}
        \toprule
        Model&GPU&Loading Memory&Token/s \\
        \midrule
        Mixtral 8$\times$7b&$2\times$A100&96.8 GB&23\\
        Mixtral 8$\times$7b&$1\times$3090&OOM&-\\
        \cdashline{1-5}
        \textbf{MC\# 2.54-bit}&$1\times$3090&16.4 GB&53\\
        \midrule

        DeepSeek-VL2-L&$1\times$A100&55.0 GB&36\\
        DeepSeek-VL2-L&$1\times$3090&OOM&-\\
        \cdashline{1-5}
        \textbf{MC\# 2.59-bit}&$1\times$3090&8.9 GB&65\\

        \bottomrule 
    \end{tabular}
    }

\end{center}
\label{tab:a_deploy}
% \vspace{-0.1in}
\end{table}

\subsubsection{Inference Latency}

In Tab.~\ref{tab:a_deploy}, we report the empirical deployment speedups achieved by our MC method across diverse hardware platforms. The gains in \textbf{MC\#} stem from static compression during the \emph{PMQ} phase and from CUDA-kernel adaptations (based on HQQ) together with \emph{OTP}. All throughput measurements were obtained with an input sequence length of 2048 tokens and an output length of 512 tokens. On an RTX 3090 GPU, MC-MoE with aggressive compression attains an average generation rate of 52 tokens/s, offering a cost-efficient deployment. In this setting, the compressed MoE LLM surpasses a dense LLM in memory footprint, accuracy, and speed.

\section{Conclusion}
MoE represents a promising framework of sparse models for multimodal understanding through scaling up the model capacity. However, the memory demands and redundancy among experts pose significant challenges for their practical implementation. In this work, we propose \textbf{MC\#}, a mixture compression strategy based on the imbalance of significance among experts. This method co-designs the \emph{Pre-Loading Mixed-Precision Quantization~(PMQ)} and \emph{Online Top-any Pruning~(OTP)} approach, allowing MoE models to be compressed to an ultra-low bit-width while maintaining exceptional memory and parameter efficiency, as well as knowledgeable performance. And our mixed-precision strategy is orthogonal to various quantization techniques. Comprehensive experiments validate the effectiveness of our mixture compression, revealing that highly compressed MoE-LLMs and MoE-VLMs can even outperform equal-size full-precision dense LLMs, thereby improving the feasibility of MoE compression. Future work will focus on adapting this strategy for multimodal applications and optimizing it for specific hardware platforms.

\textbf{Acknowledgements.} This work has been supported in part by Hong Kong Research Grant Council - Early Career Scheme (Grant No. 27209621), General Research Fund Scheme (Grant No. 17202422, 17212923), Theme-based Research (Grant No. T45-701/22-R), the Innovation and Technology Fund (Mainland-Hong Kong Joint Funding Scheme, MHP/053/21), and the Shenzhen-Hong Kong-Macau Technology Research Program (SGDX20210823103537034). This research is also supported in part by National Key R\&D Program of China (2022ZD0115502), National Natural Science Foundation of China (NO. 62461160308, U23B2010), "Pioneer" and "Leading Goose" R\&D Program of Zhejiang (No. 2024C01161).

	\bibliographystyle{IEEEtran}
\small{
\bibliography{ref}
}
% \bibliographystyle{IEEEtran}
% \bibliography{IEEEabrv,ref}

% \begin{IEEEbiography}[{\includegraphics[width=1in,height=1.25in,clip,keepaspectratio]{bio/huangshaofei.jpg}}]{Shaofei Huang}
% is currently a Ph.D. candidate at Institute of Information Engineering, Chinese Academy of Sciences, supervised by Prof. Si Liu and Prof. Jizhong Han. She received her B.S. degree from Peking University. She has published several papers on CVPR, ECCV, TIP, etc. Her research interests include image and video segmentation, and road element perception for autonomous driving.
% \end{IEEEbiography}

% \begin{IEEEbiography}[{\includegraphics[width=1in,height=1.25in,clip,keepaspectratio]{bio/shenzhenwei.jpg}}]{Zhenwei Shen} is currently an algorithm engineer at TuSimple. He received his B.S. degree in Vehicle Engineering from Shandong University of Science and Technology, Qingdao, China, in 2014, and his M.E. degree in Transportation Engineering from Shandong University of Science and Technology, Qingdao, China, in 2018. His research interests include computer vision and deep learning.
% \end{IEEEbiography}

% \begin{IEEEbiography}[{\includegraphics[width=1in,height=1.25in,clip,keepaspectratio]{bio/huangzehao.jpg}}]{Zehao Huang}
% is currently an algorithm engineer at TuSimple. He received his B.S. degree in automatic control from Beihang University, Beijing, China, in 2015. His research interests include computer vision and image processing.
% \end{IEEEbiography}

\begin{IEEEbiography}[{\includegraphics[width=1in,height=1.25in,clip,keepaspectratio]{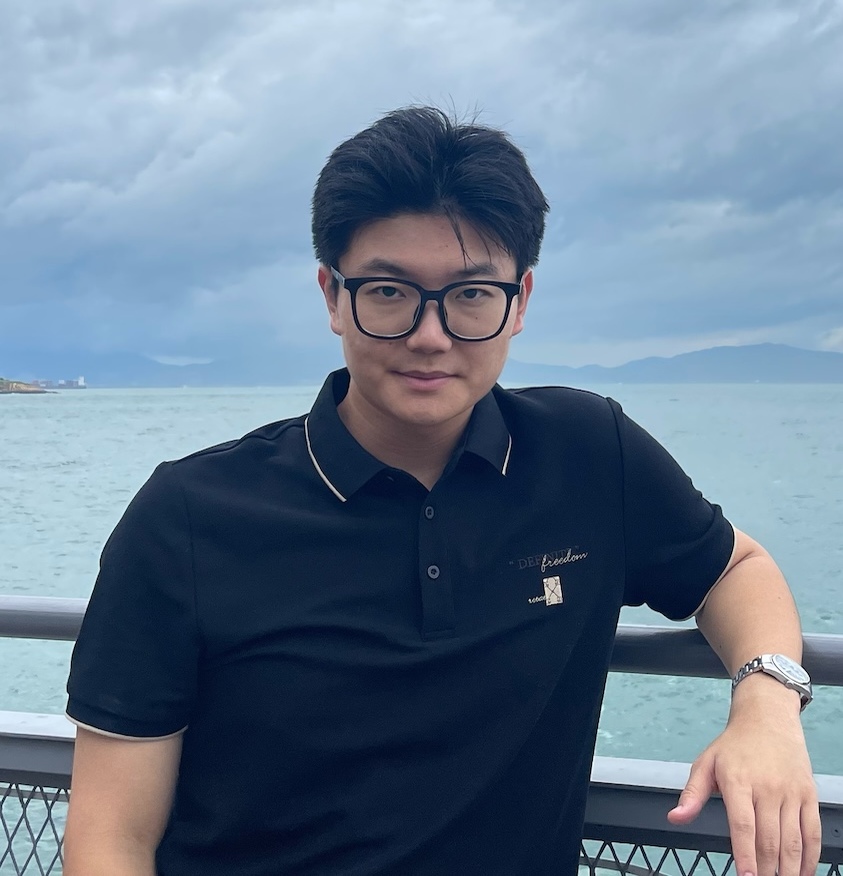}}]{Wei Huang} 
(Graduate Student Member, IEEE) is currently a Ph.D. at the Department of Electrical and Electronic Engineering (EEE) of the University of Hong Kong, as a member of the Computer Vision and Machine Intelligence (CVMI) Lab and Wearable, Intelligent and Soft Electronics (WISE) Lab. He obtained his B.S. in computer science from the School of Computer Science and Engineering, Beihang University.  His research interests include efficient multimodal LLMs, multi-modality reasoning and wearable AI. He has published several papers on top journals and conferences, including ICML, ICLR, CVPR, etc.
\end{IEEEbiography}

\begin{IEEEbiography}[{\includegraphics[width=1in,height=1.25in,clip,keepaspectratio]{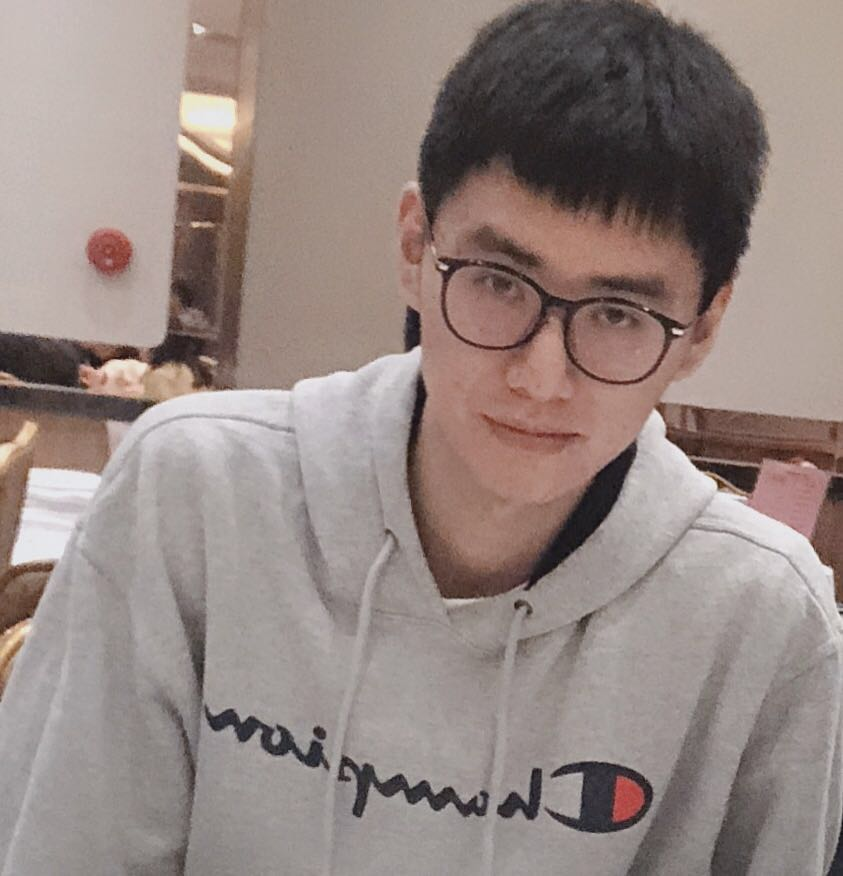}}]{Yue Liao} is currently a research fellow at School of Computing, National University of Singapore. He received his PhD degree at School of Computer Science and Engineering, Beihang University.  His research interests include multimodal understanding and embodied AI. He has published more than 30 papers at top journals and conferences, including T-PAMI, T-IP, NIPS, ICLR, CVPR, ICCV and ECCV, etc.
\end{IEEEbiography}

\begin{IEEEbiography}[{\includegraphics[width=1in,height=1.25in,clip,keepaspectratio]{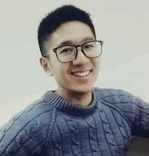}}]{Yukang Chen} is a research scientist in NVIDIA Research, working on efficient LLMs and VLMs. He received his PhD from the Chinese University of Hong Kong this year, specializing in efficient deep learning, LLMs, and computer vision. He has published over 30 papers in top conferences and journals, with 10 as the first author. His work has been featured in multiple oral presentations at top conferences such as ICLR and CVPR, and has accumulated over 5,000 citations on Google Scholar. His first-authored open-source projects have received over 6,000 stars on GitHub. Yukang was selected as a finalist candidate for the ByteDance Fellowship and has achieved winner/1st positions in renowned competitions and leaderboards, including Microsoft COCO, ScanNet, and nuScenes.
\end{IEEEbiography}

\begin{IEEEbiography}[{\includegraphics[width=1in,height=1.25in,clip,keepaspectratio]{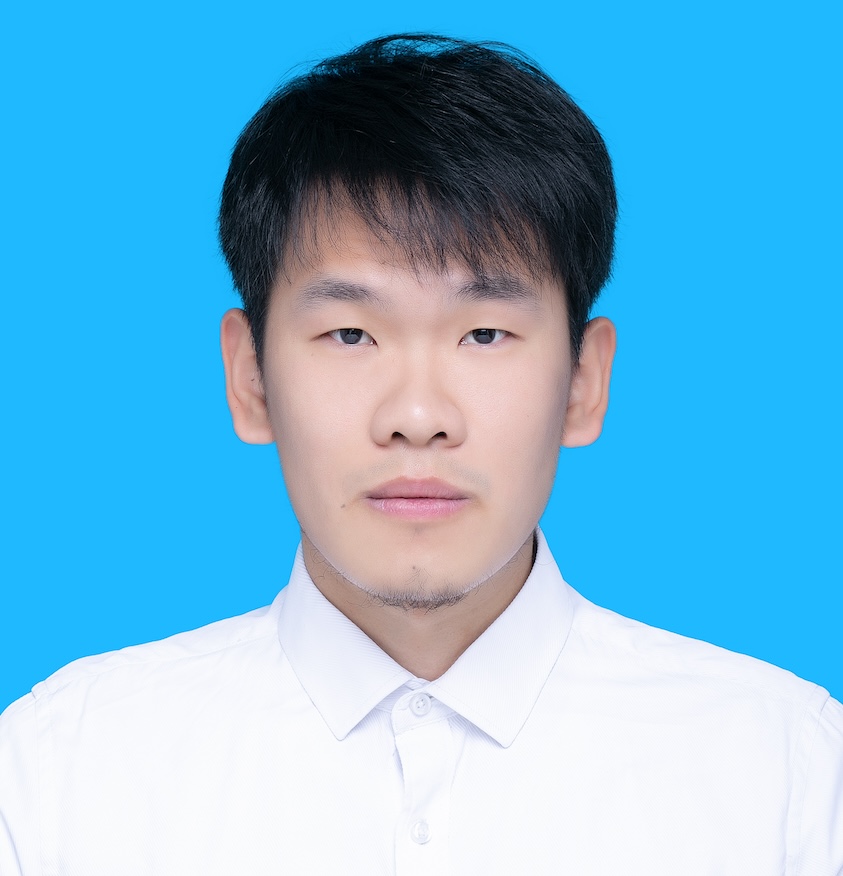}}]{Jianhui Liu} is currently a final year Ph.D student at the University of Hong Kong. He received his bachelor's degree from the School of Computer Science, Xidian University. His research interests include 3D scene understanding, 6D pose estimation and efficient model design. He has published more than 10 papers at top journals and conferences, including TKDR, NIPS, CVPR, ICCV, ICLR etc.
\end{IEEEbiography}

\begin{IEEEbiography}[{\includegraphics[width=1in,height=1.25in,clip,keepaspectratio]{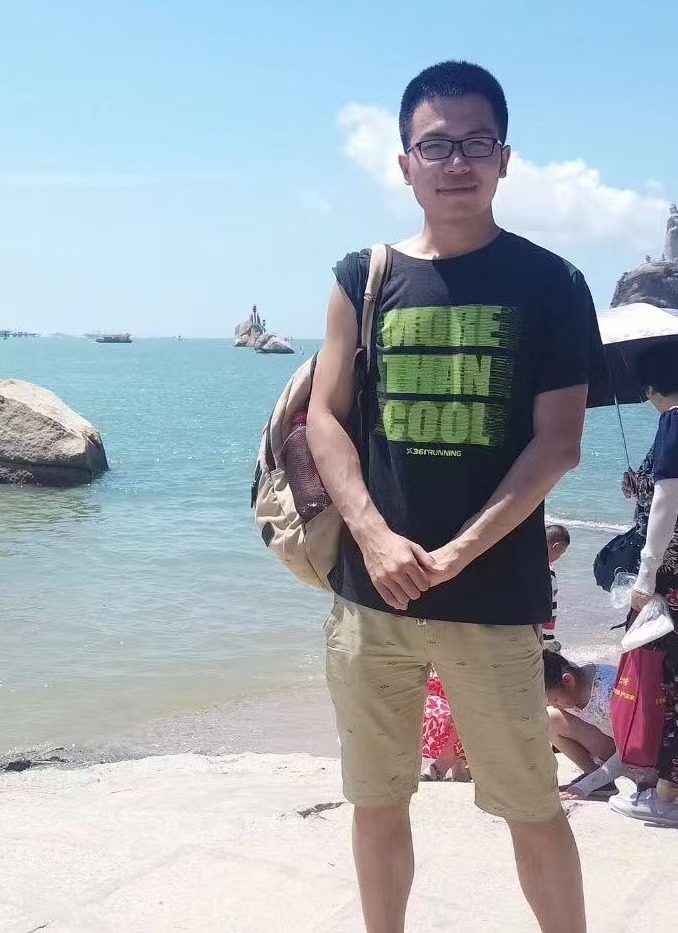}}]{Haoru Tan} is currently a second-year Ph.D student at the University of Hong Kong. He received his master's degree from the Chinese Academy of Sciences, Institute of Automation. He has also had internship experiences at institutions such as Baidu Research, Alibaba DAMO Academy, and Tencent Research. His research interests include data-centric AI, machine learning, and optimizations. He has published more than 10 papers at top journals and conferences, including NeurIPS, ICLR, ICCV, CVPR, IJCV, and T-PAMI. 
\end{IEEEbiography}

\begin{IEEEbiography}[{\includegraphics[width=1in,height=1.25in,clip,keepaspectratio]{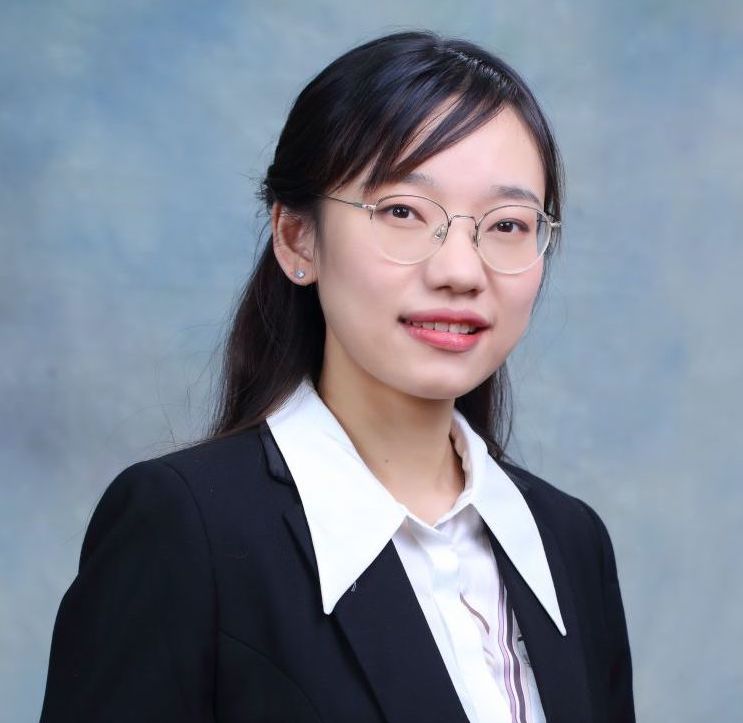}}]{Si Liu}
(Senior Member, IEEE) is currently a full professor at Institute of Artificial Intelligence, Beihang University. She is the recipient of the National Science Fund for Excellent Young Scholars. Her research interests include cross-modal multimedia intelligent analysis and classical computer vision tasks. She has published more than 60 cutting-edge papers and been cited over 13000 times on Google Scholar. She has won the Best Paper Awards of ACM MM 2021 and 2013, the Best Video Award of IJCAI 2021, and the Best Demo Award of ACM MM 2012. She is currently the associate editor of IEEE TMM, IEEE TCSVT, and CVIU, and she has served as the area chair of ICCV, CVPR, ECCV, ACM MM, and other top conferences many times.
\end{IEEEbiography}

\begin{IEEEbiography}[{\includegraphics[width=1in,height=1.25in,clip,keepaspectratio]{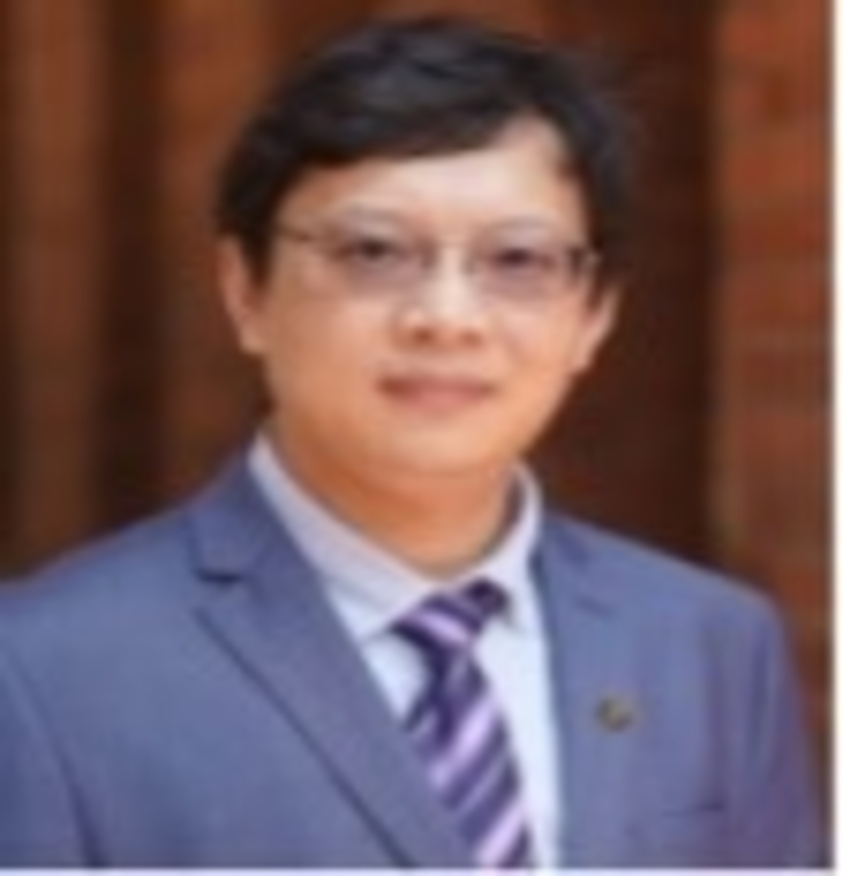}}]{Shiming Zhang}
(Member, IEEE) is currently an Assistant Professor at the Department of Electrical and Electronic Engineering (EEE) of the University of Hong Kong, leading the wearable, intelligent, and soft electronics (WISE) research group. Before that, he spent 3 years at the University of California, Los Angeles (UCLA), as a postdoctoral scholar (group leader on bioelectronics), and obtained his Ph.D. from École Polytechnique, Université de Montréal, Canada, and B.S./M.S. from Jilin University, China. He was recognized as a “Rising Star” by Advanced Materials (2024) and an “Emerging Investigator” by JMCC (2022) for his contributions to the interdisciplinary fields of soft semiconductor devices, electrochemistry, and hydrogel bioelectronics.
\end{IEEEbiography}

\begin{IEEEbiography}[{\includegraphics[width=1in,height=1.25in,clip,keepaspectratio]{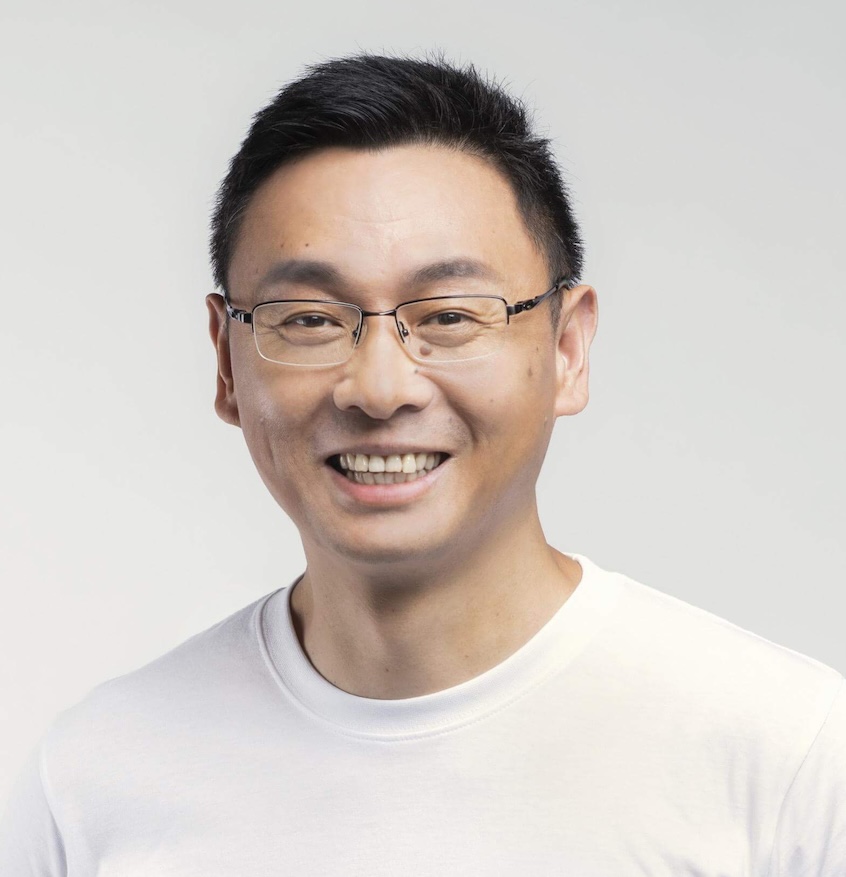}}]{Shuicheng Yan}
(Fellow, IEEE) is a Distinguished Professor (Practice) at the School of Computing, National University of Singapore, and formerly served as the Group Chief Scientist at Sea Group, alongside several other notable industry engagements. He is a Fellow of the Singapore Academy of Engineering, AAAI, ACM, IEEE, and IAPR. His research focuses on computer vision, machine learning, and multimedia analysis. To date, he has published over 800 papers in top-tier international journals and conferences, achieving an H-index exceeding 140. He has also been recognized as one of the World's Highly Cited Researchers for ten years. In addition, his team has secured ten first-place or honorable-mention accolades in two flagship competitions, Pascal VOC and ImageNet (ILSVRC). Additionally, his team has earned more than ten best paper and best student paper awards, including a remarkable grand slam at ACM Multimedia, a premier conference in multimedia research, with three Best Paper Awards, two Best Student Paper Awards, and one Best Demo Award.
\end{IEEEbiography}

\begin{IEEEbiography}[{\includegraphics[width=1in,height=1.25in,clip,keepaspectratio]{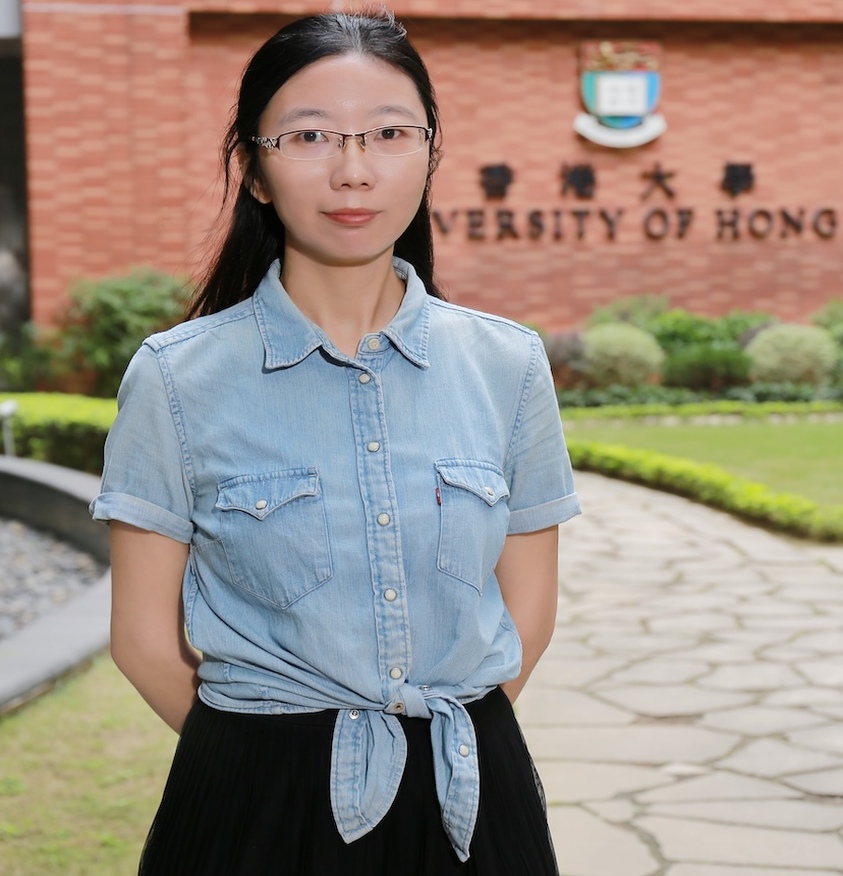}}]{Xiaojuan Qi}
(Senior Member, IEEE) is currently an assistant professor at the Department of Electrical and Electronic Engineering (EEE) of the University of Hong Kong, leading the Computer Vision and Machine Intelligence Lab (CVMI Lab). Before that, she spent 1 year at the University of Oxford, UK, as a postdoctoral scholar with Prof. Philip Torr, and obtained her Ph.D. from The Chinese University of Hong Kong, Hong Kong, and B.S. from Shanghai Jiao Tong University, China. Her research encompasses the broad areas of Computer Vision, Deep Learning, and Artificial Intelligence. She has published more than 100 cutting-edge papers and has been cited over 38000 times on Google Scholar, and has also been named among 'IEEE AI's 10 to Watch for 2024' and '35 Innovators Under 35 for China by MIT Technology. She has served as the area chair of NeurIPS, ICCV, CVPR, ECCV, and other top conferences many times.
\end{IEEEbiography}

\vfill

\end{document}